\documentclass[a4article]{clv2_arxiv}
\issue{x}{x}{xxxx}
\usepackage[a4paper]{geometry}
\usepackage{latexsym}
\usepackage{amssymb}
\usepackage{amsfonts}
\usepackage{amsmath}
\usepackage{microtype}
\usepackage{multirow}
\usepackage{graphicx}
\usepackage{booktabs}
\usepackage{subfigure}
\usepackage{tabularx}
\usepackage{fixmath}
\usepackage{color}

\runningtitle{HyperLex: A Large-Scale Evaluation of Graded Lexical Entailment}

\runningauthor{Vuli\'{c} et al.}

\begin{document}
% really have to
\title{HyperLex: A Large-Scale Evaluation of Graded Lexical Entailment}

\author{Ivan Vuli\'{c}\thanks{Language Technology Lab (LTL), Department of Theoretical and Applied Linguistics, University of Cambridge, 9 West Road, CB3 9DP Cambridge, UK. E-mail: \texttt{\{iv250|dsg40|alk23\}@cam.ac.uk}}}
\affil{LTL, University of Cambridge}

\author{Daniela Gerz\footnotemark[1]}
\affil{LTL, University of Cambridge}

\author{Douwe Kiela\thanks{Facebook AI Research, 770 Broadway, NY 10003, New York City, NY, USA. E-mail: \texttt{dkiela@fb.com}}}
\affil{Facebook AI Research}

\author{Felix Hill\thanks{Google DeepMind, 7 Pancras Square, London NC14AG, UK. E-mail: \texttt{felixhill@google.com}}}
\affil{Google DeepMind}

\author{Anna Korhonen\footnotemark[1]}
\affil{LTL, University of Cambridge}

\maketitle
\begin{abstract}
We introduce HyperLex - a dataset and evaluation resource that quantifies the extent of of the semantic category membership, that is, \textsc{type-of} relation also known as hyponymy-hypernymy or lexical entailment (LE) relation between 2,616 concept pairs. Cognitive psychology research has established that typicality and category/class membership are computed in human semantic memory as a gradual rather than binary relation. Nevertheless, most NLP research, and existing large-scale invetories of concept category membership (WordNet, DBPedia, etc.) treat category membership and LE as binary. To address this, we asked hundreds of native English speakers to indicate typicality and strength of category membership between a diverse range of concept pairs on a crowdsourcing platform. Our results confirm that category membership and LE are indeed more gradual than binary. We then compare these human judgements with the predictions of automatic systems, which reveals a huge gap between human performance and state-of-the-art LE, distributional and representation learning models, and substantial differences  between the models themselves. We discuss a pathway for improving semantic models to overcome this discrepancy, and indicate future application areas for improved graded LE systems.
\end{abstract}

\section{Introduction}
\label{s:intro}
Most native speakers of English, in almost all contexts and situations, would agree that {\em dogs}, {\em cows}, or {\em cats} are {\em animals}, and that \emph{tables} or \emph{pencils} are not. However, for certain concepts, membership of the animal category is less clear-cut. Whether  lexical concepts such as \emph{dinosaur}, \emph{human being} or \emph{amoeba} are considered animals seems to depend on the context in which such concepts are described, the perspective of the speaker or listener and even the formal scientific knowledge of the interlocutors. Despite this indeterminacy, when communicating, humans intuitively reason about such relations between concepts and categories \cite{Quillian:1967bs,Collins:1969article}. Indeed, the ability to quickly perform inference over such networks and arrive at coherent knowledge representations is crucial for human language understanding.

The Princeton WordNet lexical database \cite{Miller:1995cacm,Fellbaum:1998wn} is perhaps the best known attempt to formally represent such a semantic network. In WordNet, concepts are organised in a hierarchical fashion, in an attempt to replicate observed aspects of human semantic memory \cite{Collins:1972article,Beckwith:1991wn}. One of the fundamental relations between concepts in WordNet is the so-called {\sc type-of} or {\em hypernymy-hyponymy} relation that exists between category concepts such as \emph{animal} and their constituent members such as \emph{cat} or \emph{dog}. The type-of relation is particularly important in language understanding because it underlines the lexical entailment (LE) relation. Simply put, an instantiation of a member concept such as a cat \emph{entails} the existence of an \emph{animal}. This lexical entailment in turns governs many cases of phrasal and sentential entailment: if we know that \emph{a cat is in the garden}, we can quickly and intuitively conclude that \emph{an animal is in the garden} too.\footnote{Due to dual and inconsistent use in prior work, in this work we use the term {\em lexical entailment (LE)} in its stricter definition: it refers precisely to the taxonomical {\em hyponymy-hypernymy} relation, also known as {\sc type-of}, or {\sc is-a} relation. More details on the distinction between taxonomical and substitutable LE are provided in Sect.~\ref{s:graded}.} 

Because of this fundamental connection to language understanding, the automatic detection and modelling of lexical entailment has been an area of much focus in natural language processing \cite[inter alia]{Bos:2005emnlp,Dagan:2006pascal,Baroni:2012eacl,Beltagy:2013sem}. The ability to effectively detect and model both lexical and phrasal entailment in a human-like way may be critical for numerous related applications such as question answering, information retrieval, information extraction, and text summarisation and generation \cite{Androutsopoulos:2010jair}. For instance, in order to answer a question such as ``{\em Which mammal has a strong bite?}'', a question-answering system has to know that a jaguar or a grizzly bear are types of mammals, while a crocodile or a piranha are not.

Although inspired to some extent by theories of human semantic memory, large-scale inventories of semantic concepts, such as WordNet, typically make many simplifying assumptions, particularly regarding the nature of the \emph{type-of} relation, and consequently  the effect of LE. In WordNet, for instance, all semantic relations are represented in a binary way (i.e., concept $X$ entails $Y$) rather than gradual (e.g., $X$ entails $Y$ to a certain degree). However, since at least the pioneering experiments of prototypes \cite{Rosch:1973:natural,Rosch:1975cognitive}, it has been known that, for a given semantic category, certain member concepts are consistently understood as more central to the category than others (even when controlling for clearly confounding factors such as frequency) \cite{Coleman:1981language,Medin:1984jep,Lakoff:1990book,Hampton:2007cogsci}. In other words, WordNet and similar resources fail to capture the fact that category membership is a gradual semantic phenomenon. This limitation of WordNet also characterises much of the LE research in NLP, as we discuss later in Sect.~\ref{s:motivation}.

To address these limitations, the present work is concerned with {\em graded lexical entailment}: the degree of the LE relation between two concepts on a continuous scale. Thanks to the availability of crowdsourcing technology, we conduct a variant of the seminal behavioural data collection by Rosch \shortcite{Rosch:1973:natural}, but on a massive scale. To do so, we introduce the idea of  graded or {\em soft} LE, and design a human rating task for $(X,Y)$ concept pairs based on the following question: {\em To what degree is X a type of Y?}. We arrive at a data set with 2,616 concept pairs, each rated by at least 10 human raters, scored by the degree to which they exhibit typicality and semantic category membership and, equivalently, LE. Using this dataset, HyperLex,\footnote{HyperLex is available online at: \hspace{0.5em} \texttt{http://people.ds.cam.ac.uk/iv250/hyperlex.html}} we investigate two questions:

\begin{itemize}
\item {\bf (Q1)} Do we observe the same effects of typicality, graded membership and graded lexical entailment in human judgements as observed by Rosch? Do humans intuitively distinguish between central and non-central members of a category/class? Do humans distinguish between full and partial membership in a class as discussed by Kamp and Partee \shortcite{Kamp:1995cog}?

\item {\bf (Q2)} Is the current LE modeling and representation methodology as applied in NLP research and technology sufficient to accurately capture graded lexical entailment automatically? What is the gap between current automatic systems and human performance in the graded LE task?
\end{itemize}

The article is structured as follows. We define and discuss graded LE in Sect.~\ref{s:graded}. In Sect.~\ref{s:motivation}, we survey benchmarking resources from the literature that pertain to semantic category membership, LE identification or evaluation, and motivate the need for a new, more expressive resource. In Sect.~\ref{s:hyperlex}, we describe the design and development of HyperLex, and outline the various semantic dimensions (such as POS usage, hypernmy levels and concreteness levels) along which these concept pairs are designed to vary. 

This allows us to address Q1 in Sect.~\ref{s:analysis}, where we present a series of qualitative analyses of the data gathered and collated into HyperLex. High inter-annotator agreement scores (pairwise and mean Spearman's $\rho$ correlations around 0.85 on the entire dataset, similar correlations on noun and verb subsets) indicate that participants found it unproblematic to rate consistently the graded LE relation for the full range of concepts. These analyses reveal that the data in HyperLex enhances, rather than contradicts or undermines the information in WordNet, in the sense that hypernymy-hyponymy pairs receive highest average ratings in HyperLex compared to all other WordNet relations. We also show that participants are able to capture the implicit asymmetry of the graded LE relation by examining ratings of $(X,Y)$ and reversed $(Y,X)$ pairs. Most importantly, our analysis shows that the effects of typicality, vagueness, and gradual nature of LE are indeed captured in human judgements. For instance, graded LE scores indicate that humans rate concepts such as {\em to talk} or {\em to speak} as more typical instances of the class {\em to communicate} than concepts such as {\em to touch}, or {\em to pray}.

%FIND BETTER EXAMPLES HERE  - WRESTLING OR SOFTBALL NOT GREAT - MY EXAMPLES MAY NOT BE IN THE DATASET. VERB EXAMPLES GOOD. MAKE SURE THE EXAMPLES CONTRADICT THE FREQUENCY THING. 

In Sect.~\ref{s:experiments} we then turn our attention to Q2: we evaluate the performance of a wide range of LE detection or measurement approaches. This review covers: (i) distributional models relying on the distributional inclusion hypothesis \cite{Geffet:2005acl,Lenci:2012sem} and semantic generality computations \cite{Santus:2014eacl}, (ii) multi-modal approaches \cite{Kiela:2015acl}, (iii) WordNet-based approaches \cite{Pedersen:2004aaai}, (iv) a selection of state-of-the-art recent word embeddings, some optimised for similarity on semantic similarity data sets \cite[inter alia]{Mikolov:2013nips,Levy:2014acl,Wieting:2015tacl}, others developed to better capture the asymmetric LE relation \cite{Vilnis:2015iclr,Vendrov:2016iclr}. Due to its size, and unlike other word pair scoring data sets such as SimLex-999 or WordSim-353, in HyperLex we provide standard train/dev/test splits (both {\em random} and {\em lexical} \cite{Levy:2015naacl,Shwartz:2016arxiv}) so that HyperLex can be used for supervised learning. We therefore evaluate several prominent supervised LE architectures \cite[inter alia]{Baroni:2012eacl,Weeds:2014coling,Roller:2014coling}. Although we observe interesting differences in the models, our findings indicate clearly that none of the currently available models or approaches accurately model the relation of graded LE reflected in human subjects. This study therefore calls for new paradigms and solutions capable of capturing the gradual nature of semantic relations such as hypernymy in hierarchical semantic networks.

In Section~\ref{s:application}, we turn to the future and discuss potential applications of the graded LE concept and HyperLex. We conclude in Section~\ref{s:conclusion} by summarising the key aspects of our contribution. HyperLex offers robust, data-driven insight into how humans perceive the concepts of typicality and graded membership within the graded LE relation. We hope that this will in turn incentivise research into language technology that both reflects human semantic memory more faithfully and interprets and models linguistic entailment more effectively.

%Reasoning about semantic networks and hierarchies denoting semantic relations between concepts: essential for knowledge representation.

%Entailment as one of the fundamental concepts in logic. Because of HyperLex robots will walk the Earth in greater numbers than today, etc., etc.

\section{Graded Lexical Entailment}
\label{s:graded}
%\input{02_graded}

%\subsection{What is Graded Lexical Entailment?}
\label{ss:what}
\paragraph{Note on Terminology} 
Due to dual and inconsistent use in prior work, in this work we use the term {\em lexical entailment (LE)} in its stricter definition. It refers precisely to the taxonomical {\em hyponymy-hypernymy} relation, also known as {\sc is-a}, or {\sc type-of} relation \cite[inter alia]{Hearst:1992coling,Weeds:2004coling,Snow:2004nips,Pantel:2006acl,Do:2010emnlp}, e.g., {\em snake} is a {\sc type-of} {\em animal}, {\em computer} is a {\sc type-of} {\em machine}.

This is different from the definition used in \cite{Zhitomirsky:2009cl,Kotlerman:2010nle,Turney:2015nle} as {\em substitutable} lexical entailment: this relation holds for a pair of words $(X,Y)$ if a
possible meaning of one word (i.e., $X$) entails a meaning of the other, and the entailing word can substitute the entailed one in some typical contexts. This definition is looser and more general than the {\sc type-of} definition, as it also encompasses other lexical relations such as synonymy, metonymy, meronymy, etc.\footnote{For instance, Turney and Mohammad \shortcite{Turney:2015nle} argue that in the sentences {\em Jane dropped the glass} and {\em Jane dropped something fragile}, the concept {\em glass} should entail {\em fragile}.}

\paragraph{Definitions}
The classical definition of {\em ungraded lexical entailment} is as follows: Given a concept word pair $(X,Y)$, $Y$ is a hypernym of $X$ if and only if $X$ is a type of $Y$, or equivalently every $X$ is a $Y$.\footnote{Other variants of the same definition replace {\sc type-of} with {\sc kind-of} or {\sc instance-of}.} On the other hand, {\em graded lexical entailment} defines the strength of the lexical entailment relation between the two concepts. Given the concept pair $(X,Y)$ and the entailment strength $s$, the triplet $(X,Y,s)$ defines to what degree $Y$ is a hypernym of $X$ (i.e., {\em to what degree $X$ is a type of $Y$}), where the degree is quantified by $s$, e.g., to what degree {\em snake} is a {\sc type-of} {\em animal}.

It may be observed as approximate or {\em soft} entailment, a weaker form of the classical entailment variant \cite{Esteva:2012fuzzy,Bankova:2016arxiv}.  By imposing a threshold $thr$ on $s$, all graded relations may be straightforwardly converted to discrete ungraded decisions.

%While ungraded lexical entailment has to make a categorical (binary) decision on the nature of the relation between $X$ and $Y$, the graded or soft variant implies a continuous scale: it simply assigns the strength of entailment $s$ for the concept pair $(X,Y)$.
\begin{figure*}[t]
                        \centering
                        \subfigure[{\sc animal}]{
            \includegraphics[scale=1.02]{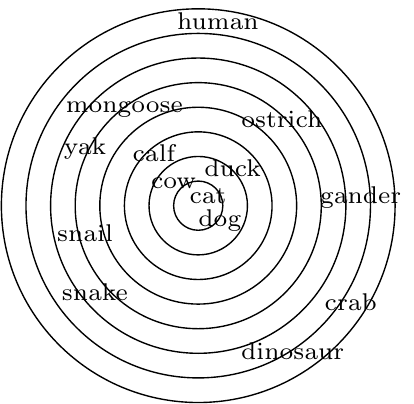}
                                \label{fig:animal}
                        }
                        \subfigure[{\sc sport}]{
             \includegraphics[scale=1.02]{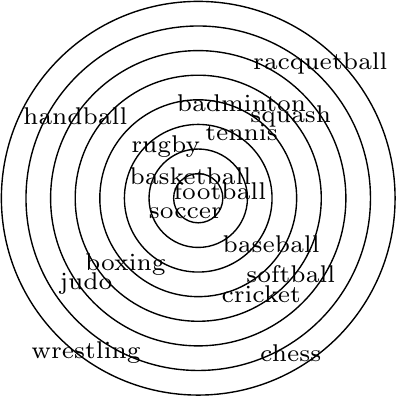}
                                \label{fig:sport}
                        }
                        \subfigure[{\sc to move}]{
             \includegraphics[scale=1.02]{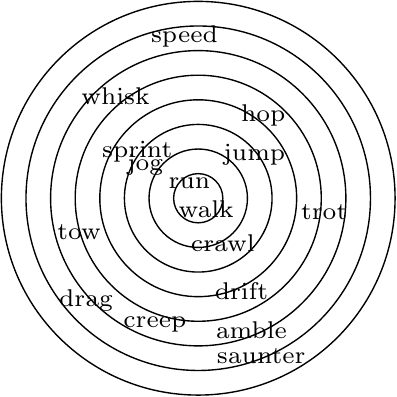}
                                \label{fig:move}
                        }
                        \vspace{-1.5mm}
                        \caption{Toy examples illustrating the ``typicality'' or centrality of various class instances $X$ of the $Y$ classes (a) {\em animal}, (b) {\em sport}, (c) {\em (to) move}.}
\vspace{-2mm}
\end{figure*}

\paragraph{(Proto)typicality, Graded Membership, and Graded LE}
The graded LE relation as described by the intuitive question ``to what degree is X a type of Y?'' encompasses two distinct phenomena described in cognitive science research (cf. \cite{Hampton:2007cogsci}). First, it can be seen as the measure of {\em typicality} in graded cognitive categorisation \cite{Rosch:1973:natural,Rosch:1975cognitive,Medin:1984jep,Lakoff:1990book}, where some instances of a category are more central than others, as illustrated in Fig.~\ref{fig:animal}-Fig.~\ref{fig:move}. It measures to what degree some class instance $X$ is a prototypical example of class/concept $Y$. For instance, when humans are asked to give an example instance of the concept {\em sport}, it turns out that {\em football} and {\em basketball} are more frequently cited than {\em wrestling}, {\em chess}, {\em softball}, or {\em racquetball}. Another viewpoint stresses that ``prototypes serve as reference points for the categorisation of not-so-clear instances'' \cite{Taylor:2003book}. Osherson and Smith \shortcite{Osherson:1997cog} make further developments to the theory of (proto)typicality by recognising that there exist concepts ``that lack prototypes while possessing degrees of exemplification''. They list the famous example of the concept \textit{building} without a clear prototype; however, people tend to agree that most banks are more typical buildings than, say, barns or pile dwellings.

Second, the graded LE relation also arises when one asks about the applicability of concepts to objects: the boundaries between a category and its instances are much more often fuzzy and vague than it is unambiguous and clear-cut \cite{Kamp:1995cog}. In other words, the \textit{graded membership} (often termed \textit{vagueness}) measures the graded applicability of a concept to different instances, e.g., it is not clear to what extent different objects in our surroundings (e.g., \textit{tables}, \textit{pavements}, \textit{washing machines}, \textit{stairs}, \textit{benches}) could be considered members of the category \textit{chair} despite the fact that such objects can be used as ``objects on which one can sit''. 

The notions of typicality and graded membership are not limited to concrete or nominal concepts, as similar gradience effects are detected for more complex and abstract concepts (e.g., {\em ``To what degree is THESIS an instance/type of STATEMENT?''}) \cite{Coleman:1981language}, or action verbs \cite{Pulman:1983book} and adjectives \cite{Dirven:1986book}.

In short, graded membership or vagueness quantifies ``whether or not and to what degree an instance falls within a conceptual category'' , while typicality reflects ``how representative an exemplar is of a category'' \cite{Hampton:2007cogsci}. The subtle distinction between the two is discussed and debated at length from the philosophical and psychological perspective \cite{Osherson:1981cog,Kamp:1995cog,Osherson:1997cog,Hampton:2006,Hampton:2007cogsci,Blutner:2013,Decock:2014nous}. In our crowdsourcing study with non-expert workers, we have deliberately avoided any explicit differentiation between the two phenomena captured by the same intuitive 'to-what-degree' question, reducing the complexity of the study design and allowing for their free variance in the collected data in terms of their quantity and representative concept pairs. In addition, the distinction is often not evident for verb concepts. We leave further developments with respect to the two related phenomena of typicality and vagueness for future work, and refer the interested reader to the aforementioned literature.

\paragraph{Relation to Relational Similarity} A strand of related research on relational similarity \cite{Turney:2006cl,Jurgens:2012semeval} also attempts to assign the score $s$ to a pair of concepts $(X,Y)$. Note that there exists a fundamental difference between relational similarity and graded lexical entailment. In the latter, $s$ refers to the degree of the LE relation in the $(X,Y)$ pair, that is, to the levels of typicality and graded membership of the instance $X$ for the class $Y$, while the former quantifies the typicality of the pair $(X,Y)$ for some fixed lexical relation class $R$ \cite{Bejar:1991book,Vylomova:2016acl}, e.g., to what degree the pair {\em (snake, animal)} reflects a typical LE relation or a typical synonymy relation.\footnote{For instance, given the lexical relation classification scheme of Bejar et al. \shortcite{Bejar:1991book}, LE or {\sc Class-Inclusion} is only one of the 10 high-level relation classes.}

\paragraph{Graded LE vs. Semantic Similarity}
A plethora of current evaluations in NLP and representation learning almost exclusively focus on semantic similarity and relatedness. Semantic similarity as quantified by e.g. SimLex-999 or SimVerb-3500 \cite{Gerz:2016emnlp} may be redefined as {\em graded synonymy relation}. The graded scores there, in fact, refer to the strength of the synonymy relation between any pair of concepts $(X,Y)$. One could say that semantic similarity aims to answer the question {\em to what degree $X$ and $Y$ are similar}.\footnote{From the SimLex-999 guidelines: ``Two words are syonymys if they have very similar meanings. Synonyms represent the same type or category (...) you are asked to compare word pairs and to rate how {\em similar} they are...'' Synonymy and LE capture different aspects of meaning regarding semantic hierarchies/taxonomies: e.g., while the pair {\em (mouse, rat)} receives a score of 7.78 in SimLex-999 (on the scale 0-10), the same pair has a graded LE score of 2.22 in HyperLex.} Therefore, an analogy between previously annotated semantic similarity data sets and our objective to construct a graded LE data set may be utilised to introduce the graded LE task and facilitate the construction of HyperLex. %Additionally, it may be used to build the first comprehensive wide-coverage evaluation set for graded lexical entailment.

\section{Design Motivation}
\label{s:motivation}
\subsection{Lexical Entailment Evaluations in NLP}
Since the work in NLP and human language understanding focuses on the ungraded version of the LE relation, we briefly survey main ungraded LE evaluation protocols in Sect.~\ref{ss:evalprot}, followed by an overview of benchmarking LE evaluation sets in Sect.~\ref{ss:sets}. We show that none of the existing evaluation protocols coupled with existing evaluation sets enables a satisfactory evaluation of the capability of statistical models to capture graded LE. As opposed to existing evaluation sets, by collecting human judgements through a crowdsourcing study our new HyperLex evaluation set also enables qualitative linguistic analysis on how humans perceive and rate graded lexical entailment.

\subsubsection{Evaluation Protocols} 
\label{ss:evalprot}
Evaluation protocols for the lexical entailment or type-of relation in NLP, based on the classical definition of ungraded LE, may be roughly clustered as follows:

\paragraph{(i) Entailment Directionality} 
Given two words $(X,Y)$ that are known to stand in a lexical entailment relation, the system has to predict the relation directionality, that is, which word is the hypernym and which word is the hyponym. More formally, the following mapping is defined by the directionality function $f_{dir}$:

\vspace{-0.9em}
{\small
\begin{align}
f_{dir}: (X,Y) \rightarrow \{-1,1\}
\end{align}}%
$f_{dir}$ simply maps to $1$ when $Y$ is the hypernym, and to $-1$ otherwise. 

\paragraph{(ii) Entailment Detection} 
The system has to predict whether there exists a lexical entailment relation between two words, or the words stand in some other relation, e.g., synonymy, meronymy-holonymy, causality, no relation, see \cite{Hendrickx:2010semeval,Jurgens:2012semeval,Vylomova:2016acl} for a more detailed overview of lexical relations. The following mapping is defined by the detection function $f_{det}$:

\vspace{-0.9em}
{\small
\begin{align}
f_{det}: (X,Y) \rightarrow \{0,1\}
\end{align}}%
$f_{det}$ simply maps to $1$ when $(X,Y)$ stand in a lexical entailment relation, irrespective to the actual directionality of the relation, and to $0$ otherwise. 

\paragraph{(iii) Entailment Detection and Directionality} 
This recently proposed evaluation protocol \cite{Weeds:2014coling,Kiela:2015acl} combines (i) and (ii). The system first has to detect whether there exists a lexical entailment relation between two words $(X,Y)$, and then, if the relation holds, it has to predict its directionality, i.e., the correct hypernym. The following mapping is defined by the joint detection and directionality function $f_{det+dir}$:

\vspace{-0.9em}
{\small
\begin{align}
f_{det+dir}: (X,Y) \rightarrow \{-1,0,1\}
\end{align}}%
$f_{det+dir}$ maps to $1$ when $(X,Y)$ stand in a lexical entailment relation and $Y$ is the hypernym, to $-1$ if $X$ is the hypernym, and to $0$ if $X$ and $Y$ stand in some other lexical relation or no relation.

\paragraph{Standard Modeling Approaches}
These decisions are typically based on the {\em distributional inclusion hypothesis} \cite{Geffet:2005acl} or a measure of {\em lexical generality} \cite{Herbelot:2013acl}. The intuition supporting the former is that the class (i.e., {\em extension}) denoted by a hyponym is included in the class denoted by the hypernym, and therefore hyponyms are expected to occur in a subset of the contexts of their hypernyms. The intuition supporting the latter hints that typical characteristics constituting the {\em intension} (i.e., concept) expressed by a hypernym (e.g., {\em move} or {\em eat} for the concept word {\em animal}) are semantically more general than the characteristics forming the
intension\footnote{The terms intension and extension assume classical intensional and extensional definitions of a concept, e.g., \cite{VanBenthem:1996book,Baronett:2012book}.} of its hyponyms (e.g., {\em bark} or {\em has tail} for the concept word {\em dog}). In other words, superordinate concepts such as {\em animal} or {\em appliance} are semantically less informative than their hyponyms \cite{Murphy:2003book}, which is also reflected in less specific contexts for hypernyms. 

Unsupervised (distributional) models of lexical entailment were instigated by the early work of Hearst \shortcite{Hearst:1992coling} on prototypicality patterns (e.g., the pattern ``X such as Y'' indicates that Y is a hyponym of X). The current unsupervised models typically replace the symmetric cosine similarity measure which works well for semantic similarity computations \cite{Bullinaria:2007brm,Mikolov:2013iclr} with an asymmetric similarity measure optimised for entailment \cite{Weeds:2004coling,Clarke:2009gems,Kotlerman:2010nle,Lenci:2012sem,Herbelot:2013acl,Santus:2014eacl}.

Supervised models, on the other hand, attempt to learn the asymmetric operator from a training set, differing mostly in the feature selection to represent each candidate pair of words \cite{Baroni:2012eacl,Fu:2014acl,Rimell:2014eacl,Weeds:2014coling,Roller:2014coling,Fu:2015taslp,Shwartz:2016arxiv,Roller:2016arxiv}.\footnote{Typical choices are feature vector concatenation ($\vec{X} \oplus \vec{Y}$), difference ($\vec{Y} - \vec{X}$), or element-wise multiplication ($\vec{X} \odot \vec{Y}$) where $\vec{X}$ and $\vec{Y}$ are feature vectors of concepts X and Y.} An overview of the supervised techniques also discussing their main shortcomings is provided by Levy et al. \shortcite{Levy:2015naacl}, while a thorough discussion of differences between unsupervised and supervised entailment models is provided by Turney and Mohammad \shortcite{Turney:2015nle}.

\paragraph{Why is HyperLex Different?} In short, regardless of the chosen methodology, the evaluation protocols (directionality or detection) may be straightforwardly translated into binary decision problems: (1) distinguishing between hypernyms and hyponyms, (2) distinguishing between lexical entailment and other relations. 

HyperLex, on the other hand, targets a different type of evaluation. The graded entailment function $f_{graded}$ defines the following mapping:

\vspace{-0.9em}
{\small
\begin{align}
f_{graded}: (X,Y) \rightarrow \mathbb{R}_0^+
\end{align}}%
$f_{graded}$ outputs the strength of the lexical entailment relation $s \in \mathbb{R}_0^+$.

By adopting the graded LE paradigm, HyperLex thus measures the degree of lexical entailment between words $X$ and $Y$ constituting the order-sensitive pair $(X,Y)$. From another perspective, it measures the typicality and graded membership of the instance $X$ for the class/category $Y$. From the relational similarity viewpoint \cite{Turney:2006cl,Jurgens:2012semeval,Zhila:2013naacl}, it also measures the prototypicality of the pair $(X,Y)$ for the LE relation. %(see Sect.~\ref{ss:what}). 

\subsubsection{Evaluation Sets}
\label{ss:sets}
\paragraph{BLESS}
Introduced by Baroni and Lenci \shortcite{Baroni:2011gems}, the original BLESS evaluation set includes 200 concrete English nouns as target concepts (i.e., $X$-s from the pairs $(X,Y)$), equally divided between animate and inanimate entities. 175 concepts were extracted from the McRae feature norms dataset \cite{McRae:2005brm}, while the remaining 25 were selected manually by the authors. These concepts were then paired to 8,625 different relatums (i.e., $Y$-s) yielding a total of 26,554 $(X,Y)$ pairs, where 14,440 contain a meaningful lexical relation and 12,154 are paired randomly. The lexical relations represented in BLESS are lexical entailment, co-hyponymy, meronymy, attribute, event, and random/no relation.

The use of its hyponymy-hypernymy/LE subset of 1,337 $(X,Y)$ pairs is then twofold. First, for directionality evaluations \cite{Santus:2014eacl,Kiela:2015acl}, only the LE subset is used. Note that original BLESS data is always presented with the hyponym first, so gold annotations are implicitly provided here. Second, for detection evaluations \cite{Santus:2014eacl,Roller:2014coling,Levy:2015naacl}, the pairs from the LE subset are taken as positive pairs, while all the remaining pairs are considered negative pairs. That way, the evaluation data effectively measures a model's ability to predict the positive LE relation. Another evaluation dataset based on BLESS was introduced by Santus et al. \shortcite{Santus:2015ws}. Following the standard annotation scheme, it comprises 7,429 noun pairs in total, and 1,880 pairs LE pairs in particular, covering a wider range of relations than BLESS (i.e., the dataset now includes synonymy and antonymy pairs).
\begin{table}[t]
\centering
\begin{footnotesize}
\def\arraystretch{0.97}
\begin{tabularx}{1.0\linewidth}{XXc}
{Variant} & {Pair} & {Annotation} \\
\toprule
\toprule
{BLESS} & {(cat, animal)} & {1} \\
\midrule
{} & {(cat, animal)} & {1} \\
{WBLESS} & {(cat, monkey)} & {0} \\
{} & {(animal, cat)} & {0} \\
\midrule
{} & {(cat, animal)} & {1} \\
{BiBLESS} & {(cat, monkey)} & {0} \\
{} & {(animal, cat)} & {-1} \\
%\bottomrule
\end{tabularx}
\vspace{-0.3em}
\caption{Example pairs from BLESS dataset variants.}
\label{tab:bless}
\end{footnotesize}
\vspace{-1.1em}
\end{table}

Adaptations of the original BLESS evaluation set were proposed recently. First, relying on its LE subset, Weeds et al. \shortcite{Weeds:2014coling} created another dataset called WBLESS \cite{Kiela:2015acl} consisting of 1,976 concept pairs in total. Only $(X,Y)$ pairs where $Y$ is the hypernym are annotated as positive examples. It also contains reversed LE pairs (i.e., $X$ is the hypernym), cohyponymy pairs, meronymy-holonymy pairs and randomly matched nouns balanced across different lexical relations, all are annotated as negative examples. Due to its construction, WBLESS is used solely for experiments on LE detection. Weeds et al. \shortcite{Weeds:2014coling} created another dataset in a similar fashion, consisting of 5,835 noun pairs, targeting co-hyponymy detection.

For the combined detection and directionality evaluation, a variant evaluation set called BiBLESS was proposed \cite{Kiela:2015acl}. It is built on WBLESS, but now explicitly distinguishes direction in LE pairs. Examples of concept pairs in all BLESS variants can be found in Tab.~\ref{tab:bless}. A majority of alternative ungraded LE evaluation sets briefly discussed here have a structure very similar to BLESS and its variants. 

\paragraph{Kotlerman et al. (2010)} Based on the original dataset of \cite{Zhitomirsky:2009cl}, this evaluation set \cite{Kotlerman:2010nle} contains 3,772 word pairs in total. The structure is similar to BLESS: 1,068 pairs are labeled as positive examples (i.e., 1 or {\em entails} iff $X$ entails $Y$), and 2,704 labeled as negative examples, including the reversed positive pairs. The assignment of binary labels is described in detail by \cite{Zhitomirsky:2009cl}. The class sizes are not balanced, and due to its design, although each pair is unique, 30 high-frequent nouns occur in each pair in the dataset. Note that this dataset has been annotated according to the broader definition of substitutable LE, see Sect.~\ref{ss:what}.

\paragraph{Baroni et al. (2012)} The $N_1 \vDash N_2$ evaluation set contains 2,770 nominal concept pairs, with 1,385 pairs labeled as positive examples (i.e., 1 or {\em entails}) \cite{Baroni:2012eacl}. The remaining 1,385 pairs labeled as negatives were created by inverting the positive pairs and randomly matching concepts from the positive pairs. The pairs and annotations were extracted automatically from WordNet and then validated manually by the authors, e.g., the abstract concepts with a large number of hyponyms such as \textit{entity} or \textit{object} were removed from the pool of concepts). 

\paragraph{Levy et al. (2014)} A similar dataset for the standard LE evaluation may be extracted from manually annotated
entailment graphs of subject-verb-object tuples (i.e., propositions) \cite{Levy:2014conll}: noun LEs were extracted from entailing tuples that were identical except for one of the arguments, thus propagating the proposition-level entailment to the word level. This data set was built for the medical domain and adopts the looser definition of substitutable LE.
\begin{table}[t]
\centering
\begin{footnotesize}
\def\arraystretch{0.98}
\begin{tabularx}{1.0\linewidth}{l l X}
{Resource} & {} & {Relation}\\
\toprule
\toprule
{WordNet} & {} & {instance hypernym, hypernym} \\
{Wikidata} & {} & {subclass of, instance of} \\
{DBPedia} & {} & {type} \\
{Yago} & {} & {subclass of} \\
%\bottomrule
\end{tabularx}
\vspace{-0.2em}
\caption{Indicators of LE/hypernymy relation in structured semantic resources.}
\label{tab:kbs}
\end{footnotesize}
\vspace{-1.4em}
\end{table}

\paragraph{Custom Evaluation Sets} A plethora of relevant work on ungraded LE do not rely on established evaluation resources, but simply extract ad-hoc LE evaluation data using distant supervision from readily available semantic resources and knowledge bases such as WordNet \cite{Miller:1995cacm}, DBPedia \cite{Auer:2007iswc}, \cite{Tanon:2016www}, Yago \cite{Suchanek:2007www}, or dictionaries \cite{Gheorghita:2012lrec}. Although plenty of the custom evaluation sets are available online, there is a clear tendency to construct a new custom dataset in every subsequent paper which uses the same evaluation protocol for ungraded LE.

A standard practice \cite[inter alia]{Snow:2004nips,Snow:2006acl,Bordes:2011aaai,Riedel:2013naacl,Socher:2013nips,Weeds:2014coling,Vendrov:2016iclr,Shwartz:2016arxiv} is to extract positive and negative pairs by coupling concepts that are directly related in at least one of the resources. Only pairs standing in an unambiguous hypernymy/LE relation, according to the set of indicators from Tab.~\ref{tab:kbs}, are annotated as positive examples (i.e., again $1$ or {\em entailing}, Tab.~\ref{tab:bless}) \cite{Shwartz:2015conll}. All other pairs standing in other relations are taken as negative instances. Using related rather than random concept pairs as negative instances enables detection experiments. We adopt a similar construction principle regarding wide coverage of different lexical relations in HyperLex. This decision will support a variety of interesting analyses related to graded LE and other relations.

\paragraph{Jurgens et al. (2012)} 
Finally, the evaluation resource most similar in spirit to HyperLex is the dataset of Jurgens et al. \shortcite{Jurgens:2012semeval} (\textit{https://sites.google.com/site/semeval2012task2/}) created for measuring degrees of relational similarity. It contains 3,218 word pairs labelled with 79 types of lexical relations from the Bejar et al.'s \shortcite{Bejar:1991book} relation classification scheme. 

The dataset was constructed using two phases of crowdsourcing. First, for each of the 79 subcategories, human subjects were shown paradigmatic examples of word pairs in the given subcategory. They
were then asked to generate more pairs of the same semantic relation type. Second, for each of the 79 subcategories, other subjects were shown word pairs that were generated in the first phase, and they were asked to rate the pairs according to their degree of prototypicality for the given semantic relation type. This is different from HyperLex where all word pairs, regardless of their actual relation, were scored according to the degree of lexical entailment between them.

The Bejar et al.'s hierarchical classification system contains ten high-level categories, with five to ten subcategories each. Only one high-level category, {\sc Class-Inclusion} refers to the true relation of ungraded LE or hypernymy-hyponymy, and the scores in the data set do not reflect graded LE. The data set aims at a wide coverage of different fine-grained relations: it comprises a small sample of manually generated instances (e.g., the number of distinct pairs for the {\sc Class-Inclusion} class is 200) for each relation scored according to their prototypicality only for that particular relation. For more details concerning the construction of the evaluation set, we refer the reader to the original work. Also, for details on how to convert the dataset to an evaluation resource for substitutable LE, we refer the reader to \cite{Turney:2015nle}.

\paragraph{HyperLex: A Short Summary of Motivation}
The usefulness of these evaluation sets is evident from their wide usage in the LE literature over recent years: they helped to guide the development of semantic research focussed on taxonomical relations. However, none of the evaluation sets contains graded LE ratings. Therefore, HyperLex may be considered as a more informative data collection: it enables a new evaluation protocol focussed on gradience of the \textsc{type-of} relation rooted in cognitive science \cite{Hampton:2007cogsci}. As discussed in Sect.~\ref{ss:what}, graded annotations from HyperLex may be easily converted to ungraded annotations: HyperLex may also be used in the standard format of previous LE evaluation sets (see Tab.~\ref{tab:bless}) for detection and directionality evaluation protocols (see later in Sect.~\ref{ss:results}).

Second, a typical way to evaluate word representation quality at present is by judging the similarity of representations assigned to similar words. The most popular semantic similarity evaluation sets such as SimLex-999 or SimVerb-3500 consist of word pairs with similarity ratings produced by human annotators. HyperLex is the first resource that can be used for the intrinsic evaluation \cite{Schnabel:2015emnlp,Faruqui:2016arxiv} of LE-based vector space models \cite{Vendrov:2016iclr}, see later in Sect.~\ref{ss:orderemb}. Encouraged by high inter annotator agreement scores and evident large gaps between the human and system performance (see Sect.~\ref{s:results}), we believe that HyperLex will guide the development of a new generation of representation-learning architectures that induce hypernymy/LE-specialised word representations, as opposed to nowadays ubiquitous word representations targeting exclusively semantic similarity and/or relatedness (see later the discussion in Sect.~\ref{ss:further} and Sect.~\ref{s:application}).

Finally, HyperLex provides a wide coverage of different semantic phenomena related to LE: graded membership vs typicality (see Sect.~\ref{ss:what}), entailment depths, concreteness levels, word classes (nouns and verbs), word pairs standing in other lexical relations, etc. Besides its primary purpose as an evaluation set, such a large-scale and diverse crowdsourced semantic resource (2,616 pairs in total!) enables novel linguistic and cognitive science analyses regarding human typicality and vagueness judgments, as well as taxonomic relationships (discussed in Sect.~\ref{s:analysis}).

\section{The HyperLex Data Set}
\label{s:hyperlex}
\paragraph{Construction Criteria}
Hill, Reichart, and Korhonen \shortcite{Hill:2015cl} argue that comprehensive high-quality evaluation resources have to satisfy the following three criteria: 

\noindent {\em (C1) Representative}: The resource covers the full range of concepts occurring in natural language. 

\noindent {\em (C2) Clearly defined}: A clear understanding is needed of what
exactly the gold standard measures, that is, the data set has to precisely define the annotated relation, e.g., relatedness as with WordSim-353, similarity as with SimLex-999, or in this case {\em graded lexical entailment}. 

\noindent (C3) {\em Consistent and reliable}: Untrained native speakers must be able to quantify the target relation consistently relying on simple instructions.

The choice of word pairs and construction of the evaluation set were steered by the requirements. The criterion C1 was satisfied by sampling a sufficient number of pairs from the University of Southern Florida (USF) Norms data set \cite{Nelson:2004usf}. As shown in prior work \cite{Hill:2015cl}, the USF data set provides an excellent range of different semantic relations (e.g., synonyms vs hypernyms vs meronyms vs cohyponyms) and semantic phenomena (e.g., it contains concrete vs abstract word pairs, noun pairs vs verb pairs). This, in turn, guarantees a wide coverage of distinct semantic phenomena in HyperLex. We discuss USF and the choice of concept words in more detail in Sect.~\ref{ss:choice}.

C2-C3 were satisfied in HyperLex by providing clear and precise annotation guidelines which accurately outline the lexical entailment relation and its graded variant in terms of the synonymous definition based on the {\sc type-of} relationship \cite{Fromking:2013book} for average native speakers of English without any linguistic background. We discuss the annotation guidelines and questionnaire structure in Sect.~\ref{ss:guidelines} and Sect.~\ref{ss:questionnaire}. 

\paragraph{Final Output}
The HyperLex evaluation set contains noun pairs (2,163 pairs) and verb pairs (453 pairs) annotated for the strength of the lexical entailment relation between the words in each pair. Since the LE relation is asymmetric and the score always quantifies to what degree $X$ is a type of $Y$, pairs $(X,Y)$ and $(Y,X)$ are considered distinct pairs. Each concept pair is rated by at least 10 human raters. The rating scale goes from 0 (no type-of relationship at all) to 10 (perfect type-of relationship). Several examples from HyperLex are provided in Tab.~\ref{tab:examples}. 

\begin{table}[!t]
\begin{center}
\def\arraystretch{0.97}
\begin{footnotesize}
\begin{tabularx}{\linewidth}{X l}
 Pair & HyperLex LE Rating \\
  \toprule
  \toprule
  chemistry / science & 10.0 \\
  motorcycle / vehicle & 9.85 \\
  pistol / weapon & 9.62 \\
  to ponder / to think & 9.40 \\
  to scribble / to write & 8.18\\
  gate / door & 6.53 \\
  thesis / statement & 6.17 \\
  to overwhelm / to defeat & 4.75 \\
  shore / beach & 3.33 \\
  vehicle / motorcycle & 1.09 \\
  enemy / crocodile & 0.33 \\
  ear / head & 0.00 \\
  %\bottomrule
\end{tabularx}
\end{footnotesize}
\end{center}
\vspace{-0.4em}
\caption{Example word pairs from HyperLex. The order of words in each pair is fixed, e.g., the pair {\em chemistry / science} should be read as {\em ``Is CHEMISTRY a type of SCIENCE?''}}
\vspace{-1.2em}
\label{tab:examples}
\end{table}

In its 2,616 word pairs, HyperLex contains 1,843 distinct noun types and 392 distinct verb types. In comparison, SimLex-999 as the standard crowdsourced evaluation benchmark for representation learning architectures focused on the synonymy relation contains 751 distinct nouns and 170 verbs in its 999 word pairs. In another comparison, the LE benchmark BLESS (see Sect.~\ref{ss:sets}) contains relations where one of the words in each pair comes from the set of 200 distinct concrete noun types. 

\subsection{Choice of Concepts}
\label{ss:choice}
\paragraph{Sources: USF and WordNet} To ensure a wide coverage of a variety semantic phenomena (C1), the choice of candidate pairs is steered by two standard semantic resources available online: (1) the USF norms data set\footnote{http://w3.usf.edu/FreeAssociation/} \cite{Nelson:2004usf}, and (2) WordNet\footnote{https://wordnet.princeton.edu/} \cite{Miller:1995cacm}.

USF was used as the primary source of concept pairs. It is a large database of free association data collected for English, generated by presenting human subjects with one of $5,000$ cue concepts and asking them to write the first word coming to mind that is associated with that concept. Each cue concept $c$ was normed in this way by over 10 participants, resulting in a set of associates $a$ for each cue, for a total of over $72,000$ $(c,a)$ pairs. For each such pair, the proportion of participants who produced associate $a$ when presented with cue $c$ can be used as a proxy for the strength of association between the two words.  

The norming process guarantees that two words in a pair have a degree of semantic association which correlates well with semantic relatedness reflected in different lexical relations between words in the pairs. Inspecting the pairs manually revealed a good range of semantic relationship values represented, e.g., there were examples of ungraded LE pairs ({\em car / vehicle}, {\em biology / science}), cohyponym pairs ({\em peach / pear}), synonyms or near-synonyms ({\em foe / enemy}), meronym-holonym pairs ({\em heel / boot}), and antonym pairs ({\em peace / war}). USF also covers different POS categories: nouns ({\em winter / summer}), verbs ({\em to elect / to select}), and adjectives ({\em white / gray}), at the same time spanning word pairs at different levels of concreteness ({\em panther / cat} vs {\em wave / motion} vs {\em hobby / interest}). The rich annotations of the USF data (e.g., concreteness scores, association strength) can be combined with graded LE scores to yield additional analyses and insight.

WordNet was used to automatically assign a fine-grained lexical relation to each pair in the pool of candidates, which helped to guide the sampling process to ensure a wide coverage of word pairs standing in a variety of lexical relations \cite{Shwartz:2016arxiv}.

\paragraph{Lexical Relations} To guarantee the coverage of a wide range of semantic phenomena, we have conditioned the cohort/pool used for sampling on the lexical relation between the words in each pair. As mentioned above, the information was extracted from WordNet. We consider the following lexical relations in HyperLex:

\noindent (1) \texttt{hyp-N}: $(X,Y)$ pairs where $X$ is a hyponym of $Y$ according to WordNet. $N$ denotes the path length between the two concepts in the WordNet hierarchy, e.g., the pair {\em cathedral / building} is assigned the \texttt{hyp-3} relation. Due to unavailability of a sufficient number of pairs for longer paths, we have grouped all pairs with the path length $\geq 4$ into a single relation class \texttt{hyp$\geq$4}.

It was shown that pairs that are separated by fewer levels in the WordNet hierarchy are both more strongly associated and rated as more similar \cite{Hill:2015cl}. This fine-grained division over different LE levels will enable analyses based on the semantic distance in a concept hierarchy.

\noindent (2) \texttt{rhyp-N}: The same as \texttt{hyp-N}, now with the order reversed: $X$ is now a hypernym of $Y$. Such pairs were included to investigate the inherent asymmetry of the type-of relation and how human subjects perceive it.

\noindent (3) \texttt{cohyp}: $X$ and $Y$ are two instances of the same implicit category, that is, they share a hypernym (e.g., {\em dog} and {\em elephant} are instances of the category {\em animal}). For simplicity, we retain only $(X,Y)$ pairs that share a direct hypernym.

\noindent (4) \texttt{mero}: It denotes the {\sc Part-Whole} relation, where $X$ always refer to the meronym (i.e., {\sc Part}), and $Y$ to the holonym (i.e., {\sc Whole}): {\em finger / hand}, {\em letter / alphabet}. By its definition, this relation is observed only between nominal concepts.

\noindent (5) \texttt{syn}: $X$ and $Y$ are synonyms and near-synonyms, e.g., {\em movement / motion}, {\em attorney / lawyer}. In case of polysemous concepts, at least one sense has to be synonymous with a meaning of the other concept, e.g., {\em author / writer}.

\noindent (6) \texttt{ant}: $X$ and $Y$ are antonyms, e.g., {\em beginning / end}, {\em day / night}, {\em to unite / to divide}.

\noindent (7) \texttt{no-rel}: $X$ and $Y$ do not stand in any lexical relation, including the ones not present in HyperLex (e.g., causal relations, space-time relations), and are also not semantically related. This relation specifies that there is no apparent semantic connection between the two concepts at all, e.g., {\em chimney / swan}, {\em nun / softball}.

\paragraph{POS Category} HyperLex includes subsets of pairs from two principle meaning-bearing POS categories: nouns and verbs.\footnote{We have decided to leave out adjectives: they are represented in USF to a lesser extent than nouns and verbs, and it is thus not possible to sample large enough subsets of adjectives pairs across different lexical relations and lexical entailment levels, i.e., only \texttt{syn} and \texttt{ant} adjective pairs are available in USF.} This decision will enable finer-grained analyses based on the two main POS categories. It is further supported by recent research in distributional semantics showcasing that different word classes (e.g., nouns vs verbs) require different modeling approaches and distributional information to reach per-class peak performances \cite{Schwartz:2015conll}. In addition, we expect verbs to have fuzzier category borders due to their high variability and polysemy, increased abstractness, and a wide range of syntactic-semantic behaviour \cite{Jackendoff:1972book,Levin:1993book,Gerz:2016emnlp}.

\paragraph{Pools of Candidate Concept Pairs} The initial pools for sampling were selected as follows. First, we extracted all possible noun pairs (N / N) and verb pairs (V / V) from USF based on the associated POS tags available as part of USF annotations. Concept pairs of other and mixed POS (e.g., {\em puzzle / solve}, {\em meet / acquaintance}) were excluded from the pool of candidate pairs.\footnote{POS categories are generally considered to reflect very broad ontological classes \cite{Fellbaum:1998wn}. We thus felt it would be very difficult, or even counter-intuitive, for annotators to rate mixed POS pairs.} To ensure that semantic association between concepts in a pair is not accidental, we then discarded all such USF pairs that had been generated by two or fewer participants in the original USF experiments.\footnote{The numbers are available as part of USF annotations.} We also excluded all concept pairs containing a multi-word expression (e.g., {\em put down / insult}, {\em stress / heart attack}), pairs containing a named entity (e.g., {\em Europe / continent}), and pairs containing a potentially offensive concept (e.g., {\em weed / pot}, {\em heroin / drug}).\footnote{Note that the pairs with the same concept without any offensive connotation were included in the pools, e.g., {\em weed / grass}, {\em weed / plant}, or {\em ecstasy / feeling}.}

All remaining concept pairs were then assigned a lexical relation according to WordNet. In case of duplicate $(X,Y)$ and $(Y,X)$ pairs, only one variant (i.e., $(X,Y)$) was retained. In addition, all \texttt{rhyp-N} pairs at this stage were reversed into \texttt{hyp-N} pairs. All \texttt{no-rel} pairs from USF were also discarded at this stage to prevent the inclusion of semantically related pairs in the \texttt{no-rel} subset of HyperLex.

In the final step, all remaining pairs were divided into per-relation pools of candidate noun and verb pairs for each represented relation: \texttt{hyp-N}, \texttt{cohyp}, \texttt{mero}, \texttt{syn}, \texttt{ant}. Two additional pools were created for \texttt{rhyp-N} and \texttt{no-rel} after the sampling process.

\begin{figure*}[t]
\centering
\includegraphics[width=0.98\linewidth]{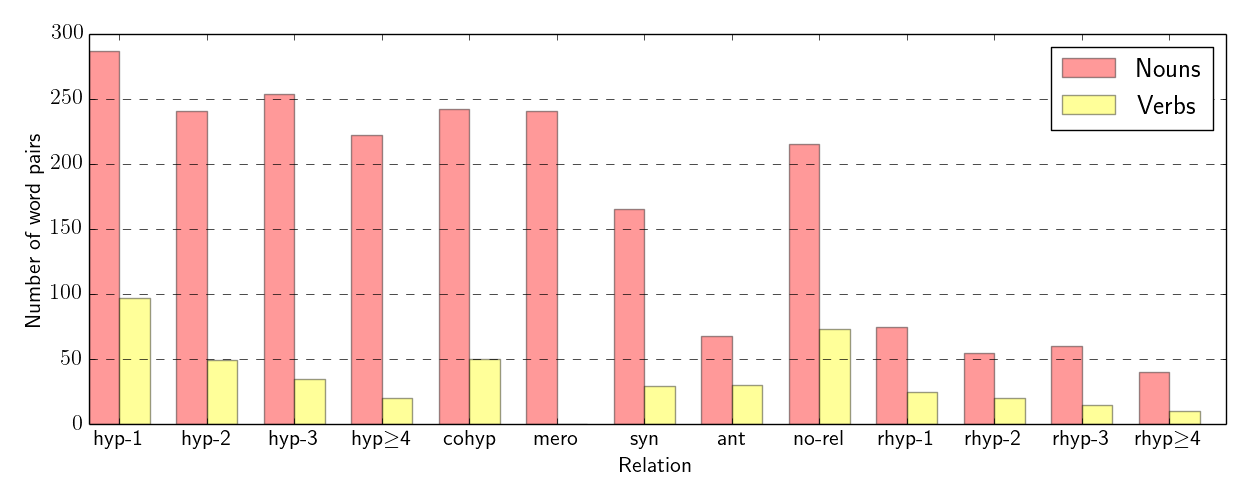}
\vspace{-1.0em}
\caption{A total number of noun and verb pairs in HyperLex representing different fine-grained semantic relations extracted from WordNet.}
\vspace{-1.6em}
\label{fig:reldistr}
\end{figure*}

\subsection{Sampling Procedure}
\label{ss:sampling}
The candidate pairs were then sampled from the respective per-relation pools. The final number of pairs per each relation and POS category was influenced by: (1) the number of candidates in each pool (therefore, HyperLex contains significantly more noun pairs), (2) the focus on LE (therefore, HyperLex contains more \texttt{hyp-N} pairs at different LE levels), (3) the wide coverage of most prominent lexical relations (therefore, each lexical relation is represented by a sufficient number of pairs), and (4) logistic reasons (we were unable to rate all candidates in a crowdsourcing study and had to sample a representative subset of candidates for each relation and POS category in the first place).

\paragraph{Step 1: Initial Sampling} 
First, pairs for lexical relations \texttt{hyp-N}, \texttt{cohyp}, \texttt{mero}, \texttt{syn}, and \texttt{ant} were sampled from their respective pools. WordNet, although arguably the best choice for our purpose, is not entirely reliable as a gold standard resource with occasional inconsistencies and debatable precision regarding the way lexical relations have been encoded: e.g., {\em silly} is a hyponym of {\em child} according to WordNet. Therefore, all sampled pairs were manually checked by the authors plus two native English speakers in several iterations. Only such sampled pairs where the majority of human checkers agreed on the lexical relation were retained. If a pair was discarded, another substitute pair was randomly sampled if available, and again verified against human judgements. 

\paragraph{Step 2: Reverse and No-Rel Pairs}
Before the next step, the pool for \texttt{rhyp-N} was generated by simply reversing the order of concepts in all previously sampled $(X,Y)$ \texttt{hyp-N} pairs. The pool for \texttt{no-rel} was generated by pairing up the concepts from the pairs extracted in Step 1 at random using the Cartesian product. From these random parings, we excluded those that coincidentally occurred elsewhere in USF (and therefore had a degree of association), as well as those that were assigned any lexical relation according to WordNet. From the remaining pairs, we accepted only those in which both concepts had been subject to the USF norming procedure, ensuring that these non-USF pairs were indeed unassociated rather than simply not normed. \texttt{rhyp-N} and \texttt{no-rel} were then sampled from these two pools, followed by another manual check. The \texttt{rhyp-N} pairs will be used to test the asymmetry of human and system judgements (see later in Tab.~\ref{tab:relations} and Tab.~\ref{tab:reversed}), which is immanent to the LE relation.

\begin{figure*}[t]
                        \centering
                        \subfigure[{Page 1}]{
           \fbox{ \includegraphics[width=0.45\linewidth]{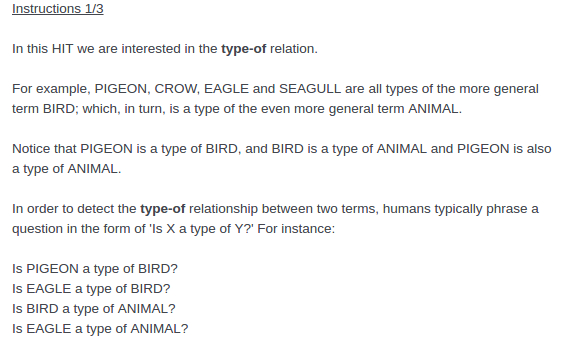}}
		\label{fig:instructions01}
                        }
                        \subfigure[{Page 2(a)}]{
            \fbox{ \includegraphics[width=0.45\linewidth]{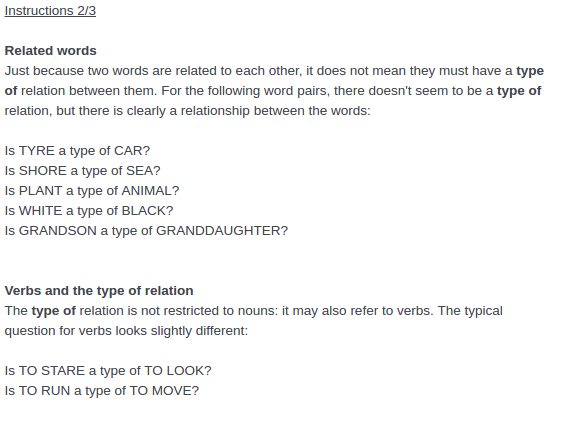}}
                                 \label{fig:instructions02a}
                        }
                        \caption{Page 1 and Page 2(a) of HyperLex annotation guidelines.}
\vspace{-1.2em}
\end{figure*}
Fig.~\ref{fig:reldistr} shows the exact numbers of noun and verb pairs across different lexical relations represented in HyperLex. The final set of 2,616 distinct word pairs\footnote{The final number was obtained after randomly discarding a small number of pairs for each relation in order to distribute the pairs of both POS categories into tranches of equal size in the crowdsourcing study, see Sect.~\ref{ss:questionnaire}.} was then annotated in a crowdsourcing study (Sect.~\ref{ss:guidelines} and Sect.~\ref{ss:questionnaire}). 

%Sampling from the USF norms annotated with WordNet relations: only the true hypernym-hyponym pairs were retained (based on the info from WordNet). The sampling process was guided by the constraints in terms of numbers of pairs for each relation that we wanted to include in the dataset: hypernymy or type-of at different levels of hierarchy (hyp: hyp-1, hyp-2, etc.), synonymy/antonymy, co-hyponymy, meronymy, no-relation. Subsets of true hyp pairs at each level were also reversed (e.g., r-hyp-1, r-hyp-2) to control for the asymmetry of the type-of relation and how humans perceive it.

\subsection{Question Design and Guidelines}
\label{ss:guidelines}
Here, we show and detail the exact annotation guidelines followed by the participants in the crowdsourcing study. In order to accurately outline the lexical entailment relation to average native speakers of English without any linguistic background, we have deliberately eschewed the usage of any expert linguistic terminology in the annotation guidelines, and have also avoided addressing the subtle differences between typicality and vagueness (Sect.~\ref{s:graded}). For instance, the terms such as {\em hyponymy/hypernymy}, {\em lexical entailment}, {\em prototypicality}, or {\em taxonomy} were never explicitly defined using any precise linguistic formalism.

\paragraph{(Page 1)} We have adopted a simpler and more intuitive definition of lexical entailment instead, based on the {\em type-of} relationship between words in question \cite{Fromking:2013book}, illustrated by a set of typical examples in the guidelines (see Fig.~\ref{fig:instructions01}). 

\paragraph{(Page 2)} Following that, a clear distinction was made between words standing in a broader relationship of {\em semantic relatedness} and words standing in an actual type-of relation (see Fig.~\ref{fig:instructions02a}). We have included typical examples of related words without any entailment relation, including meronymy pairs ({\em tyre / car}), cohyponymy pairs ({\em plant / animal}), and antonymy pairs ({\em white / black}), and pairs in other lexical relations (e.g., {\em shore / sea}). Since HyperLex also contains verbs, we have also provided several examples for a type-of relation between verbs (see again Fig.~\ref{fig:instructions02a}).

\begin{figure*}[t]
                        \centering
                        \subfigure[{Page 2(b)}]{
           \fbox{ \includegraphics[width=0.45\linewidth]{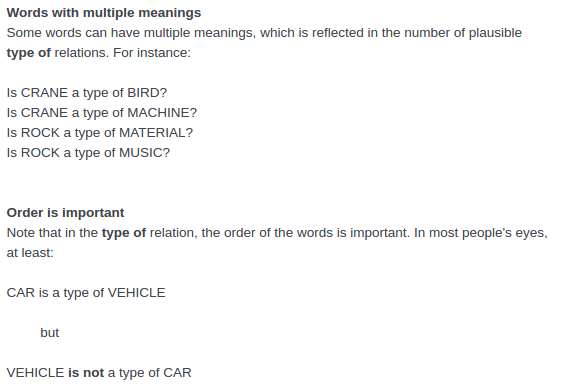}}
		\label{fig:instructions02b}
                        }
                        \subfigure[{Page 3}]{
            \fbox{ \includegraphics[width=0.45\linewidth]{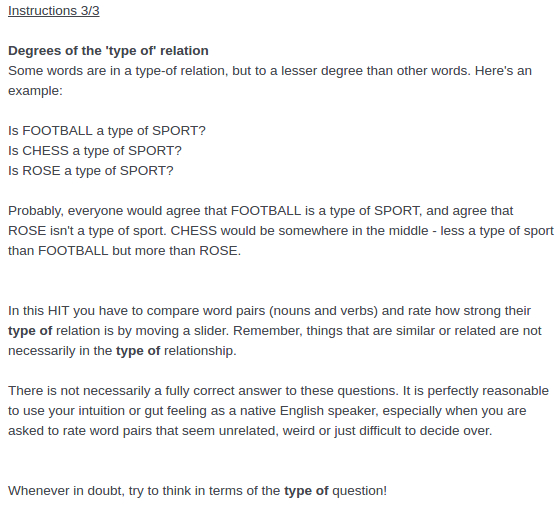}}
                                 \label{fig:instructions03}
                        }
                        \caption{Page 2(b) and Page 3 of HyperLex annotation guidelines.}
\vspace{-1.2em}
\end{figure*}

Potential polysemy issues have been addressed by stating (using intuitive examples) that {\em two words stand in a type-of relation if any of their senses stand in a type-of relation}. However, we acknowledge that this definition is vague, and the actual disambiguation process was left to the annotators and their intuition as native speakers. A similar context-free rating was used in the construction of other word pair scoring datasets such as SimLex-999 \cite{Hill:2015cl} or WordSim-353 \cite{Finkelstein:2002tois}.\footnote{Determining the set of exact senses for a given concept, and then the set of contexts that represent those senses, introduces a high degree of subjectivity into the
design process. Furthermore, in the infrequent case that some concept $X$ in a pair $(X,Y)$ is genuinely
(etymologically) polysemous, $Y$ can provide sufficient context to disambiguate $X$ \cite{Hill:2015cl,Leviant:2015arxiv}.} In the next step, we have explicitly stressed that the type-of relation is {\em asymmetric} (see Fig.~\ref{fig:instructions02b}). 

\paragraph{(Page 3)} The final page explains the main idea behind graded lexical entailment, graded membership and prototypical class instances according to the theories from cognitive science \cite{Rosch:1973:natural,Rosch:1975cognitive,Lakoff:1990book,Hampton:2007cogsci,Divjak:2013cogling} by providing another illustrative set of examples (see Fig.~\ref{fig:instructions03}). The main goals of the study were then quickly summarised in the final paragraph, and the annotators were reminded to think in terms of the type-of relationship throughout the study.

\begin{figure*}[t]
                        \centering
                        \subfigure[{Qualification question}]{
           \fbox{ \includegraphics[width=0.35\linewidth]{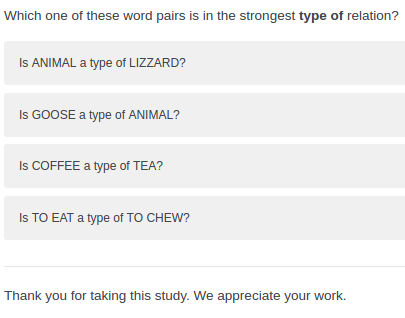}}
		\label{fig:question}
                        }
                        \subfigure[{Survey structure}]{
            \fbox{ \includegraphics[width=0.56\linewidth]{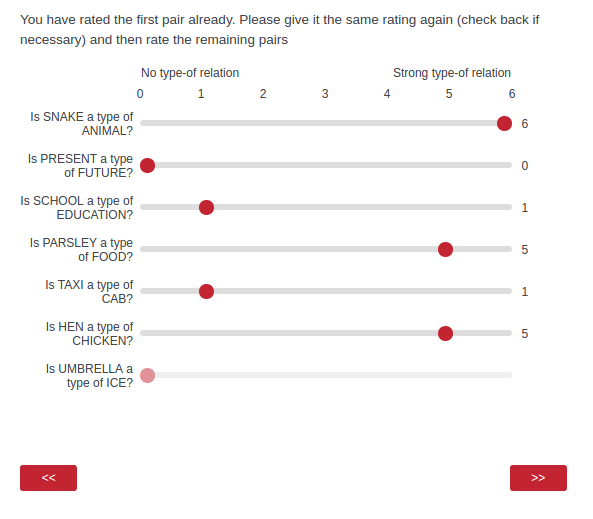}}
                                 \label{fig:survey}
                        }
                        \caption{(a) Qualification question. The correct answer is {\em ``Is GOOSE a type of ANIMAL?'}' (b) A group of noun pairs to be rated by moving the sliders. The rating slider was initially at position 0, and it was possible to attribute a rating of 0, although it was necessary to have actively moved the slider to that position to proceed to the next page. The first pair is repeated from the previous page, while the last pair will be repeated on the next page.}
\vspace{-0.8em}
\end{figure*}

\subsection{Questionnaire Structure and Participants}
\label{ss:questionnaire}
We employ the Prolific Academic (PA) crowdsourcing platform,\footnote{https://prolific.ac/  (We chose PA for logistic reasons.)} an online marketplace very similar to Amazon Mechanical Turk and to CrowdFlower. While PA was used to recruit participants, the actual questionnaire was hosted on Qualtrics.\footnote{https://www.qualtrics.com/}  Unlike other crowdsourcing platforms, PA collects and stores detailed demographic information from the participants upfront. This information was used to carefully select the pool of eligible participants. We restricted the pool to native English speakers with a 90\% approval rate (maximum rate on PA), of age 18-50, born and currently residing in the United States or the United Kingdom. 

Immediately after the guidelines, similar to the SimLex-999 questionnaire, a {\em qualification question} is posed to the participant to test whether she/he understood the guidelines and is allowed to proceed with the questionnaire. The question is: {} Fig.~\ref{fig:question}. In the case of an incorrect answer, the study terminates for the participant without collecting any ratings. 

In case of a correct answer, the participant begins rating pairs by moving a slider, as shown in Fig.~\ref{fig:survey}. Having a slider attached to the question {\em ``Is X a type of Y?''} implicitly translates the posed question to the question {\em ``To what degree is X a type of Y?''} (Sect.~\ref{ss:what}). The pairs are presented to the participant in groups of six or seven. As with SimLex-999, this group size was chosen because the (relative) rating of a set of pairs implicitly requires pairwise comparisons between all pairs in that set. Therefore, larger groups would have significantly increased the cognitive load on the annotators. Since concept pairs were presented to raters in batches defined according to POS, another advantage of grouping was the clear break (submitting a set of ratings and moving to the next page) between the tasks of rating noun and verb pairs. For better inter-group calibration, from the second group onward the last pair of the previous group became the first pair of the present group. The participants were then asked to re-assign the rating previously attributed to the first pair before rating the remaining new items (Fig.~\ref{fig:survey}).

It is also worth stressing that we have decided to retain the type-of structure of each question explicitly for all word pairs so that raters are constantly reminded of the targeted lexical relation, i.e., all $(X,Y)$ word pairs are rated according to the question {\em ``Is X a type of Y?''}, as shown in Fig.~\ref{fig:survey}. For verbs, we have decided to use the infinitive form in each question, e.g., {\em ``Is TO RUN a type of TO MOVE?''}

Following a standard practice in crowdsourced word pair scoring studies \cite{Finkelstein:2002tois,Luong:2013conll,Hill:2015cl}, each of the 2,616 concept pairs has to be assigned at least 10 ratings from 10 different accepted annotators. We collected ratings from more than 600 annotators in total. To distribute the workload, we divided the 2,616 pairs into 45 tranches, with 79 pairs each: 50 are unique to one tranche, while 20 manually chosen pairs are in all tranches to ensure consistency: the use of such consistency pairs enabled control for possible systematic
differences between annotators and tranches, which could detected by variation on
this set of 20 pairs shared across all tranches. The remaining 9 are duplicate pairs displayed to the same participant multiple times to detect inconsistent annotations. The number of noun and verb pairs is the same across all tranches (64/79 and 15/79 respectively). Each annotator was asked to rate the pairs in a single tranche only. Participants took 10 minutes on average to complete one tranche, including the time spent reading the guidelines and answering the qualification question.

\subsection{Post-Processing}
85\% of total exclusions occurred due to crowdsourcers answering the qualification question incorrectly: we did not collect any ratings from such workers. In the post-processing stage, we additionally excluded ratings of annotators who (a) did not give equal ratings to duplicate pairs; (b) showed suspicious rating patterns (e.g., randomly alternating between two ratings, using one single rating throughout the study, or assigning random ratings to pairs from the consistency set). The final acceptance rate was 85.7\% (if we also count the workers who answered the qualification question incorrectly for the total number of assignments) and 97.5\% (with such workers excluded from the counts). We then calculated the average of all ratings from the accepted raters ( $\geq$ 10 ) for each word pair. The score was finally scaled linearly from the 0-6 to the 0-10 interval as in \cite{Hill:2015cl}.

\section{Analysis}
\label{s:analysis}
\paragraph{Inter-Annotator Agreement} 
We report two different inter-annotator agreement (IAA) measures. \textbf{IAA-1 (pairwise)} computes the average pairwise Spearman's $\rho$ correlation between any two raters. This is a common choice in previous data collection in distributional semantics \cite{Pado:2007emnlp,Reisinger:2010emnlp,Silberer:2014acl,Hill:2015cl}.
\begin{table}[!t]
\begin{center}
\def\arraystretch{1.0}
\begin{footnotesize}
\begin{tabularx}{\linewidth}{X ll}
 Benchmark & IAA-1 & IAA-2 \\
  \toprule
  \toprule
  {\sc WordSim (353)} & {0.611} & {0.756} \\
  {\cite{Finkelstein:2002tois}} & {} & {} \\
  {\sc WS-Sim (203)} & {0.667} & {0.651} \\
  {\cite{Agirre:2009naacl}} & {} & {} \\
  {\sc SimLex (999)} & {0.673} & {0.778} \\
  {\cite{Hill:2015cl}} & {} & {} \\
  \midrule
  \midrule
  {\sc HyperLex (2616)} & {\bf 0.854} & {\bf 0.864} \\
  \midrule
  {\sc HyperLex: Nouns (2163)} & {0.854} & {0.864} \\
  {\sc HyperLex: Verbs (453)} & {0.855} & {0.862} \\
  %\bottomrule
\end{tabularx}
\end{footnotesize}
\end{center}
\vspace{-0.4em}
\caption{A comparison of HyperLex IAA with several prominent crowdsourced semantic similarity/relatedness evaluation benchmarks that also provide scores for word pairs. Numbers in parentheses refer to the total number of word pairs in each evaluation set.}
\vspace{-0.8em}
\label{tab:iaa}
\end{table}
\begin{table*}[t]
\begin{center}
\def\arraystretch{1.0}
\begin{scriptsize}
\begin{tabularx}{\linewidth}{l lllXXXXXlllll}
  \toprule
  {} & {hyp-1} & {hyp-2} & {hyp-3} & {hyp$\geq$4} & {cohyp} & {mero} & {syn} & {ant} & {no-rel} & {rhyp-1} & {rhyp-2} & {rhyp-3} & {rhyp$\geq$4} \\
  \cmidrule(lr){2-14}
  {IAA-1} & {0.850} & {0.844} & {0.859} & {0.848} & {0.857} & {0.856} & {0.860} & {0.858} & {0.854} & {0.855} & {0.842} & {0.868} & {0.856} \\ 
  {IAA-2} & {0.866} & {0.847} & {0.872} & {0.851} & {0.875} & {0.876} & {0.883} & {0.858} & {0.859} & {0.845} & {0.850} & {0.846} & {0.859} \\
  %\bottomrule
\end{tabularx}
\end{scriptsize}
\end{center}
\vspace{-0.3em}
\caption{Inter-annotator agreements, measured by average pairwise Spearman's $\rho$ correlation over different fine-grained semantic relations extracted from WordNet.}
\vspace{-0.8em}
\label{tab:iaarelations}
\end{table*}

A complementary measure would smooth individual annotator effects. For this aim, our \textbf{IAA-2 (mean)} measure compares the average correlation of a human rater with the average of all the other raters.  It arguably serves as better 'upper bound' than IAA-1 for the performance of automatic systems. HyperLex obtains $\rho$ = 0.854 (IAA-1) and $\rho$ = 0.864 (IAA-2), a very good agreement compared to other prominent crowdsourced benchmarks for semantic evaluation which also used word pair scoring (see Tab.~\ref{tab:iaa}).\footnote{Note that the IAAs are not computed on the entire data set, but are in fact computed per tranche, as one worker annotated only one tranche. Exactly the same IAA computation was used previously by Hill et al. \shortcite{Hill:2015cl}.} We also report IAAs over different groups of pairs according to the relation extracted from WordNet in Tab.~\ref{tab:iaarelations}. 

We acknowledge the fact that the grading process at places requires specific world-knowledge (e.g., {\em to what degree is SNAKE a type of REPTILE?}, {\em to what degree is TOMATO a type of FRUIT?}), or is simply subjective and demographically biased (e.g., {\em to what degree is TO PRAY a type of TO COMMUNICATE?}), which needs principled qualitative analyses. However, the HyperLex inter-rater agreement scores suggest that participants were able to understand the characterisation of graded lexical entailment presented in the instructions and to apply it to concepts of various types (e.g. nouns vs verbs, concrete vs abstract concepts, different lexical relations from WordNet) consistently. 

\paragraph{Typicality in Human Judgements} 
In the first qualitative analysis, we aim to investigate the straightforward question: are some concepts really more (proto)typical of semantically broader higher-level classes? Several examples of prominent high-level taxonomical categories along with LE scores are shown in Tab.~\ref{tab:graded_examples}. We might draw several preliminary insights based on the presented lists. There is an evident prototyping effect present in human judgements: concepts such as {\em cat}, {\em monkey} or {\em cow} are more typical instances of the class {\em animal} than the more peculiar instances such as {\em mongoose} or {\em snail} according to HyperLex annotators. Instances of the class {\em sport} also seem to be sorted accordingly, as higher scores are assigned to arguably more prototypical sports such as {\em basketball}, {\em volleyball} or {\em soccer}, and less protypical sports such as {\em racquetball} or {\em wrestling} are assigned lower scores.
\begin{table*}[t]
\centering
\begin{scriptsize}
\def\arraystretch{1.0}
\begin{tabularx}{1.0\linewidth}{lX lX lX lX lX lX}
\toprule
{\sc type-of} & {\bf animal} & {} & {\bf food} & {} & {\bf plant} & {} & {\bf sport} & {} & {\bf person} & {} & {\bf vehicle} \\
\cmidrule(lr){1-2} \cmidrule(lr){3-4} \cmidrule(lr){5-6} \cmidrule(lr){7-8} \cmidrule(lr){9-10} \cmidrule(lr){11-12} 
{cat} & {10.0} & {sandwich} & {10.0}  & {rose} & {9.75}  & {basketball} & {10.0}  & {girl} & {9.85}  & {car} & {10.0} \\
{monkey} & {10.0} & {pizza} & {10.0}  & {cactus} & {9.58}  & {hockey} & {10.0}  & {customer} & {9.08}  & {limousine} & {10.0} \\
{cow} & {10.0} & {rice} & {10.0}  & {flower} & {9.45}  & {volleyball} & {10.0}  & {clerk} & {8.97}  & {motorcycle} & {9.85} \\
{bat} & {9.52} & {hamburger} & {9.75}  & {lily} & {9.40}  & {soccer} & {9.87}  & {citizen} & {8.63}  & {van} & {9.75} \\
{mink} & {9.17} & {mushroom} & {9.07}  & {weed} & {9.23}  & {baseball} & {9.75}  & {nomad} & {8.63}  & {automobile} & {9.58} \\
{snake} & {8.75} & {pastry} & {8.83}  & {orchid} & {9.08}  & {softball} & {9.55}  & {poet} & {7.78}  & {tractor} & {9.37} \\
{snail} & {8.62} & {clam} & {8.20}  & {ivy} & {9.00}  & {cricket} & {9.37}  & {guest} & {7.22}  & {truck} & {9.23} \\
{mongoose} & {8.33} & {snack} & {7.78}  & {tree} & {8.63}  & {racquetball} & {9.03}  & {mayor} & {6.67}  & {caravan} & {8.33} \\
{dinosaur} & {8.20} & {oregano} & {5.97}  & {clove} & {8.47}  & {wrestling} & {8.85}  & {publisher} & {6.03}  & {buggy} & {8.20}  \\
{crab} & {7.27} & {rabbit} & {5.83}  & {turnip} & {8.05}  & {recreation} & {2.46}  & {climber} & {5.00}  & {bicycle} & {8.00} \\
{plant} & {0.13} & {dinner} & {4.85}  & {fungus} & {4.75}  & {-} & {-}  & {idol} & {4.28}  & {vessel} & {6.38} \\
%\bottomrule
\end{tabularx}
\vspace{-0.2em}
\caption{Graded LE scores for instances of several prominent taxonomical categories/classes represented in HyperLex (i.e., the categories are the word $Y$ in each $(X,Y,s)$ graded LE triplet).}
\label{tab:graded_examples}
\end{scriptsize}
\vspace{-1.2em}
\end{table*}

Nonetheless, the majority of \texttt{hyp-N} pairs $(X, animal)$ or $(X, sport)$, where $X$ is a hyponym of {\em animal/sport} according to WN, are indeed assigned reasonably high graded LE scores. It suggests that humans are able to: (1) judge the LE relation consistently and decide that a concept indeed stands in a type-of relation with another concept, and (2) grade the LE relation by assigning more strength to more prototypical class instances. Similar patterns are visible with other class instances from Tab.~\ref{tab:graded_examples}, as well as with other prominent nominal classes (e.g., {\em bird}, {\em appliance}, {\em science}). We also observe the same effect with verbs, e.g., {\em (drift, move, 8.58), (hustle, move, 7.67), (tow, move, 7.37), (wag, move, 6.80), (unload, move, 6.22)}.

We have also analysed if the effects of graded membership/vagueness (see the discussion in Sect.~\ref{ss:what}) are also captured in the ratings, and our preliminary qualitative analysis suggests so. For instance, an interesting example quantifies the graded membership in the class \textit{group}: {\em (gang, group, 9.25), (legion, group, 7.67), (conference, group, 6.80), (squad, group, 8.33), (caravan, group, 5.00), (grove, group, 3.58), (herd, group, 9.23), (fraternity, group, 8.72), (staff, group, 6.28)}. 

\begin{table}[!t]
\begin{center}
\def\arraystretch{0.95}
\begin{footnotesize}
\begin{tabularx}{\linewidth}{l XXX}
 \toprule
 {} & {\bf All} & {\bf Nouns} & {\bf Verbs} \\
 \cmidrule(lr){2-4}
 {hyp-1} & {7.86} & {7.99} & {7.49} \\
 {hyp-2} & {8.10} & {8.31} & {7.08} \\
 {hyp-3} & {8.16} & {8.39} & {6.55} \\
 {hyp$\geq$4} & {8.33} & {8.62} & {5.12} \\
 {cohyp} & {3.54} & {3.29} & {4.76} \\
 {mero} & {3.14} & {3.14} & {-} \\
 {syn} & {6.83} & {6.69} & {7.66} \\
 {ant} & {1.47} & {1.57} & {1.25} \\
 {no-rel} & {0.85} & {0.64} & {1.48} \\
 {rhyp-1} & {4.75} & {4.17} & {6.45} \\
 {rhyp-2} & {4.19} & {3.44} & {6.15} \\
 {rhyp-3} & {3.07} & {2.72} & {4.47} \\
 {rhyp$\geq$4} & {2.85} & {2.54} & {4.11} \\
  %\bottomrule
\end{tabularx}
\end{footnotesize}
\end{center}
\vspace{-0.3em}
\caption{Average HyperLex scores across all pairs, and noun and verb pairs representing finer-grained semantic relations extracted from WordNet.}
\vspace{-1.2em}
\label{tab:relations}
\end{table}

\paragraph{Hypernymy/LE Levels} Graded LE scores in HyperLex averaged for each WordNet relation are provided in Tab.~\ref{tab:relations}. Note that the LE level is extracted as the shortest direct path between two concept words in the WordNet taxonomy, where $X$-s in each $(X,Y)$ pair always refer to the less general concept (i.e., hyponym). The scores suggest several important observations. 

Graded LE scores for nouns increase with the increase of the LE level (i.e., WN path length) between the concepts. A longer WN path implies a clear difference in semantic generality between nominal concepts which seems to be positively correlated with the degree of the LE relation and ease of human judgement. A similar finding in directionality and detection experiments on BLESS and its variants was reported by Kiela et al. \shortcite{Kiela:2015acl}. They demonstrate that their model is less accurate on concepts with short paths (i.e., the lowest results are reported for WN \texttt{hyp-1} pairs from BLESS), and the performance increases with the increase of the WN path length. The tendency is explained by the lower difference in generality between concepts with short paths, which may be difficult to discern for a statistical model. The results from Tab.~\ref{tab:relations} show that human raters also display a similar tendency when rating nominal pairs. 

Another factor underlying the observed scores might be the link between HyperLex and the source USF norms. Since USF contains free association norms, one might assume that more prototypical instances are generated more frequently as responses to cue words in the original USF experiments. This, in turn, reflects in their greater presence in HyperLex, especially for concept pairs with longer WN distances.

Further, nominal concepts higher in the WN hierarchy typically refer to semantically very broad but well-defined categories such as {\em animal}, {\em food}, {\em vehicle}, or {\em appliance} (see again Tab.~\ref{tab:graded_examples}). Semantically more specific instances of such concepts are easier to judge as {\em true} hyponyms (using the ungraded LE terminology), which also reflects in higher LE ratings for such instances. However, gradience effects are clearly visible even for pairs with longer WN distances (Tab.~\ref{tab:graded_examples}). 

The behaviour with respect to the LE level is reversed for verbs: the average  scores decrease over increasing LE levels. We may attribute this effect to a higher level of abstractness and ambiguity present in verb concepts higher in the WN hierarchy stemming from a fundamental cognitive difference: Gentner \shortcite{Gentner:2006} showed that children find verb concepts harder to learn than noun concepts, and Markman and Wisniewski \shortcite{Markman:1997jep} present evidence that different cognitive operations are used when comparing two nouns or two verbs .For instance, it is intuitive to assume that human subjects find it easier to grade instances of the class {\em animal} than instances of verb classes such as {\em to get}, {\em to set} or {\em to think}.

\paragraph{LE Directionality}
Another immediate analysis investigates whether the inherent asymmetry of the type-of relation is captured by the human annotations in HyperLex. Several illustrative example pairs and their reverse pairs split across different LE levels are shown in Tab.~\ref{tab:reversed}. Two important conclusions may be drawn from the analysis.
\begin{table*}[t]
\centering
\begin{scriptsize}
\def\arraystretch{1.0}
\begin{tabularx}{1.0\linewidth}{Xll Xll Xll}
\toprule
\multicolumn{3}{X}{{hyp-1} vs {rhyp-1} (89\%)} & \multicolumn{3}{X}{{hyp-2} vs {rhyp-2} (95\%)} & \multicolumn{3}{X}{{hyp-3} vs {rhyp-3} (96\%)} \\
\cmidrule(lr){1-3} \cmidrule(lr){4-6} \cmidrule(lr){7-9} 
{Pair} & {
\texttt{scr}} & {\texttt{rscr}} & {Pair} & {\texttt{scr}} & {\texttt{rscr}} & {Pair} & {\texttt{scr}} & {\texttt{rscr}} \\
\cmidrule(lr){1-3} \cmidrule(lr){4-6} \cmidrule(lr){7-9} 
(computer, machine) & {\bf 9.83} & {2.43} & (gravity, force) & {\bf 9.50} & {3.58} & (flask, container) & {\bf 9.37} & {1.83} \\
(road, highway) & {\bf 9.67} & {4.30} & (professional, expert) & {\bf 6.37} & {6.03} & (elbow, joint) & {\bf 7.18} & {1.07}\\
(dictator, ruler) & {\bf 9.87} & {6.22} & (therapy, treatment) & {\bf 9.17} & {4.10} & (nylon, material) & {\bf 9.75} & {1.42}\\
(truce, peace) & {\bf 8.00} & {6.38} & (encyclopedia, book) & {\bf 8.93} & {2.22} & (choir, group) & {\bf 8.72} & {2.43}\\
(remorse, repentance) & {\bf 7.63} & {3.50} & (empathy, feeling) & {\bf 8.85} & {2.42} & (beer, beverage) & {\bf 9.25} & {0.67}\\
(disagreement, conflict) & {\bf 8.78} & {8.67} & (shovel, tool) & {\bf 9.70} & {2.57} & (reptile, animal) & {\bf 9.87} & {1.17}\\
(navigator, explorer) & {6.80} & {\bf 7.63}  & (fraud, deception) & {\bf 9.52} & {8.17} & (parent, ancestor) & {\bf 7.00} & {6.17}\\
(ring, jewelry) & {\bf 10.0} & {2.78} & (bed, furniture) & {\bf 9.75} & {2.63} & (note, message) & {\bf 9.00} & {6.07} \\
(solution, mixture) & {6.52} & {\bf 7.37} & (verdict, judgment) & {\bf 9.67} & {7.57} & (oven, appliance) & {\bf 9.83} & {1.33} \\
(spinach, vegetable) & {\bf 10.0} & {0.55} & (reader, person) & {\bf 7.43} & {3.33} & (king, leader) & {\bf 8.67} & {4.55}\\
(surgeon, doctor) & {\bf 8.63} & {4.05} & (vision, perception) & {3.82} & {\bf 6.25} & (hobby, activity) & {\bf 7.12} & {6.83}\\
(hint, suggestion) & {\bf 8.75} & {7.03} & (daughter, child) & {\bf 9.37} & {2.78} & (prism, shape) & {\bf 7.50} & {2.70}\\
%\bottomrule
\end{tabularx}
\vspace{-0.3em}
\caption{A selection of scored $(X,Y)$ word pairs from HyperLex holding the \texttt{hyp-1}, \texttt{hyp-2}, and \texttt{hyp-3} relation according to WordNet along with the HyperLex score for the actual $(X,Y)$ pair ({$scr$}) and the HyperLex score for the reversed $(Y,X)$ pair (i.e., \texttt{rhyp-N} relations): $rscr$. The reported percentages on top refer to the ratio of $(X,Y)$ pairs for each relation where $scr>rscr$.}
\label{tab:reversed}
\end{scriptsize}
\vspace{-1.2em}
\end{table*}

First, human raters are able to capture the asymmetry as the strong majority of \texttt{hyp-N} pairs is rated higher than their \texttt{rhyp-N} counterparts: 94\% of all \texttt{hyp-N} pairs for which exists the \texttt{rhyp-N} counterpart are assigned a higher rating. Second, the ability to clearly detect the correct LE direction seems to increase with the increase of semantic distance in WordNet: (1) We notice decreasing average scores for the \texttt{rhyp-N} relation as N increases (see Tab.~\ref{tab:relations}), (2) We notice a higher proportion of \texttt{hyp-N} concept pairs scoring higher than their \texttt{rhyp-N} counterparts as N increases (see Tab.~\ref{tab:reversed}). There are evident difficulties to decide on the entailment direction with several pairs (e.g., {\em navigator / explorer}, {\em solution / mixture}, {\em disagreement / conflict}), especially for the taxonomically closer \texttt{hyp-1} pairs, a finding aligned with prior work on LE directionality \cite{Rimell:2014eacl,Kiela:2015acl}.

\paragraph{Other Lexical Relations}
Another look into Tab.~\ref{tab:relations}, where graded LE scores are averaged across each WN-based lexical relation, indicates the expected order of all other lexical relations sorted by the average per-relation scores (i.e., \texttt{syn} $>$ \texttt{cohyp} $>$ \texttt{mero} $>$ \texttt{ant} $>$ \texttt{no-rel}). \texttt{no-rel} and \texttt{ant} pairs have the lowest graded LE scores by a large margin, as expected. \texttt{no-rel} pairs are expected to have completely non-overlapping semantic fields, which facilitates human judgement. With antonyms, the graded LE question may be implicitly reformulated as {\em To what degree is X a type of $\neg$X?} (e.g., {\em winner / loser, to depart / to arrive}), which intuitively should result in low graded LE scores: the HyperLex ratings confirm the intuition.

Low scores for \texttt{cohyp} pairs in comparison to \texttt{hyp-N} pairs indicate that the annotators are able to effectively distinguish between the two related but different well-defined taxonomical relations (i.e., \texttt{hyp-N} vs \texttt{cohyp}). High scores for \texttt{syn} pairs are also aligned with our expectations and agree with intuitions from prior work on ungraded LE \cite{Rei:2014conll}. In a slightly simplified view, given that two synonyms may be observed as two different utterances of the same semantic concept $X$, the graded LE question may be rephrased as {\em To what degree is X a type of X?}. One might say that \texttt{syn} could be seen as a special case: the degenerate taxonomical \texttt{hyp-0} relation. Such an implicit reformulation of the posed question naturally results in higher scores for \texttt{syn} pairs on average.
\begin{table}[!t]
\begin{center}
\def\arraystretch{0.95}
\begin{footnotesize}
\begin{tabularx}{\linewidth}{l XXXX}
 \toprule
 {} & {G1} & {G2} & {G3} & {G4} \\
 \cmidrule(lr){2-5}
 {\# Pairs} & {979} & {259} & {883} & {344} \\
  %\bottomrule
\end{tabularx}
\end{footnotesize}
\end{center}
\vspace{-0.3em}
\caption{A total number of concept pairs in each of the four coarse-grained groups based on the concepts' concreteness ratings: both concepts are concrete (USF concreteness rating $\geq$ 4) $\rightarrow$ $G_1$; both abstract $\rightarrow$ $G_2$; one concrete and one abstract concept with a difference in concreteness $\leq$ 1 $\rightarrow$ $G_3$ or $>$ 1 $\rightarrow$ $G_4$. \texttt{rhyp-N} pairs are not counted.}
\vspace{-1.2em}
\label{tab:concreteness}
\end{table}

\paragraph{Concreteness}
Differences in human and computational concept learning and representation have been attributed to the effects of concreteness, the extent to which a concept has a directly perceptible physical referent \cite{Paivio:1991,Hill:2014cogsci}. Since the main focus of this work is not on the distinction between abstract and concrete concepts, we have not explicitly controlled for the balanced amount of concrete/abstract pairs in HyperLex. However, since the source USF dataset provides concreteness scores, we believe that HyperLex will also enable various additional analyses regarding this dimension in future work.

Here, we report the number of pairs in four different groups based on concreteness ratings of two concepts in each pair. The four groups are as follows: ($G_1$) both concepts are concrete (USF concreteness rating $\geq$ 4); ($G_2$) both concepts are abstract (USF rating $<$ 4), ($G_3$) one concept is considered concrete and another abstract, with their difference in ratings $\leq 1$, ($G_4$) one concept is considered concrete and another abstract, with their difference in ratings $>1$.

The statistics regarding HyperLex pairs divided into groups $G_1-G_4$ is presented in Tab.~\ref{tab:concreteness}. \texttt{rhyp-N} pairs are not counted as they are simply reversed \texttt{hyp-N} pairs present in HyperLex. Concept pairs where at least one concreteness rating is missing in the USF data are also not taken into account. Although HyperLex contains more concrete pairs overall, there is also a large sample of highly abstract pairs and mixed pairs. For instance, HyperLex contains 125 highly abstract concept pairs, with both concepts scoring $\leq$ 3 in concreteness, e.g., {\em misery / sorrow, hypothesis / idea, competence / ability}, or {\em religion / belief}. This preliminary coarse-grained analysis already hints that HyperLex provides a good representation of concepts across the entire concreteness scale. This will also facilitate further analyses related to concept concreteness and its influence on the automatic construction of semantic taxonomies.

\paragraph{Data Splits: Random and Lexical}
A common problem in scored/graded word pair datasets is the lack of a standard split to development and test sets \cite{Faruqui:2016arxiv}. Custom splits, e.g., 10-fold cross-validation make results incomparable with others. Further, due to their limited size, they also do not support supervised learning, and do not provide splits into training, development, and test data. The lack of standard splits in such word pair datasets stems mostly from small size and poor coverage -- issues which we have solved with HyperLex.
\begin{table*}[!t]
\begin{center}
\def\arraystretch{0.99}
\begin{scriptsize}
\begin{tabularx}{\linewidth}{l lXXXXX}
 \toprule
 {} & \multicolumn{5}{l}{\bf Rating Intervals} \\
 \cmidrule(lr){2-7}
 {Split} & {All} & {$[0,2>$} & {$[2,4>$} & {$[4,6>$} & {$[6,8>$} & {$[8,10]$} \\
 \cmidrule(lr){2-7}
 {\bf HyperLex-All} & {2616 (2163 + 453)} & {604 (504 + 100)} & {350 (304 + 46)} & {307 (243 + 64)} & {515 (364 + 151)} & {840 (748 + 92)} \\
 {\bf Random Split} \\
 {\sc Train} & {1831 (1514 + 317)} & {423 (353 + 70)} & {245 (213 + 32)} & {215 (170 + 45)} & {361 (255 + 106)} & {587 (523 + 64)} \\
 {\sc Dev} & {130 (108 + 22)} & {30 (25 + 5)} & {17 (15 + 2)} & {15 (13 + 2)} & {26 (18 + 8)} & {42 (37 + 5)}\\
 {\sc Test} & {655 (541 + 114)} & {151 (126 + 25)} & {88 (76 + 12)} & {77 (60 + 17)} & {128 (91 + 37)} & {211 (188 + 23)} \\
  {\bf Lexical Split} \\
 {\sc Train} & {1133 (982 + 151)} & {253 (220 + 33)} & {140 (122 + 18)} & {129 (109 + 20)} & {195 (148 + 47)} & {416 (383 + 33)} \\
 {\sc Dev} & {85 (71 + 14)} & {20 (18 + 2)} & {13 (11 + 2)} & {11 (8 + 3)} & {17 (10 + 7)} & {24 (24 + 0)} \\ 
 {\sc Test} & {269 (198 + 71)} & {65 (52 + 13)} & {37 (29 + 8)} & {41 (31 + 10)} & {63 (37 + 26)} & {63 (49 + 14)} \\
  %\bottomrule
\end{tabularx}
\end{scriptsize}
\end{center}
\vspace{-0.3em}
\caption{HyperLex data splits: Basic statistics. The number of pairs is always provided in the following format \#Overall (\#N + \#V), where \#N and \#V denote the number of noun and verb pairs respectively. The columns represent groups/buckets of word pairs according to their LE scores.}
\vspace{-1.2em}
\label{tab:stats}
\end{table*}

We provide two standard data splits into train, dev, and test data: {\em random} and {\em lexical}. In the random split, 70\% of all pairs were reserved for training, 5\% for development, and 25\% for testing. The subsets were selected by random sampling, but controlling for a broad coverage in terms of similarity ranges, i.e., non-similar and highly similar pairs, as well as pairs of medium similarity are represented. Some statistics are available in Tab.~\ref{tab:stats}. A manual inspection of the subsets revealed that a good range of lexical relations is represented in the subsets.

The lexical split, advocated in \cite{Levy:2015naacl,Shwartz:2016arxiv}, prevents the effect of ''lexical memorisation'': supervised distributional lexical inference models tend to learn an independent property of a single concept in the pair instead of learning a relation between the two concepts.\footnote{For instance, if the training set contains concept pairs {\em (dog / animal)}, {\em (cow, animal)}, and {\em (cat, animal)}, all assigned very high LE scores or annotated as positive examples in case of ungraded LE evaluation, the algorithm may learn that {\em animal} is a prototypical hypernym, assigning any new $(X, animal)$ pair a very high score, regardless of the actual relation between $X$ and {\em animal}; additional analyses provided in Sect.~\ref{ss:regression}.} To prevent such behaviour, we split HyperLex into a train and test set with zero lexical overlap. We tried to retain roughly the same 70\%/25\%/5\% ratio in the lexical split. Note that the lexical split discards all ``cross-set'' training-test concept pairs. Therefore, the number of instances in each subset is lower than with the random split. Statistics are again given in Tab.~\ref{tab:stats}.

We believe that the provided standardised HyperLex data splits will enable easy and direct comparisons of various LE modeling architectures in unsupervised and supervised settings. Following arguments from prior work, we hold that it is important to provide both data set splits, as they can provide additional possibility to assess differences between models. It is true that training a model on a lexically split dataset may result in a more general model \cite{Levy:2015naacl}, which is able to better reason over pairs consisting of two unseen concepts during inference. However, Shwartz, Goldberg, and Dagan \shortcite{Shwartz:2016arxiv} argue that a random split emulates a more typical ``real-life'' reasoning scenario, where inference involves an unseen concept pair $(X,Y)$, in which $X$ and/or $Y$ have already been observed separately. Models trained on a random split may introduce the model with a concept's ``prior belief'' of being a frequent hypernym or a hyponym. This information can be effectively exploited during inference.

\section{Evaluation Setup and Models}
\label{s:experiments}
\paragraph{Evaluation Setup} We compare the performance of prominent models and frameworks focused on modeling lexical entailment on our new HyperLex evaluation set now measuring the strength of the lexical entailment relation. Due to the evident similarity of the graded evaluation with standard protocols in the semantic similarity (i.e., synonymy detection) literature \cite[inter alia]{Finkelstein:2002tois,Agirre:2009naacl,Hill:2015cl,Schwartz:2015conll}, we adopt the same evaluation setup. Each evaluated model assigns a score to each pair of words measuring the strength of lexical entailment relation between them.\footnote{Note that, unlike with similarity scores, the score now refers to an asymmetric relation stemming from the question {\em ``Is X a type of Y''} for the word pair $(X,Y)$. Therefore, the scores for two reverse pairs $(X,Y)$ and $(Y,X)$ should be different, see also Tab.~\ref{tab:reversed}.}

As in prior work on intrinsic semantic evaluations with word pair scoring evaluation sets, e.g., \cite{Hill:2015cl,Levy:2015tacl} as well as on measuring relational similarity \cite{Jurgens:2012semeval}, all reported scores are Spearman's $\rho$ correlations between the ranks derived from the scores of the evaluated models and the human scores provided in HyperLex. In this work, we evaluate off-the-shelf unsupervised models and insightful baselines on the entire HyperLex. We also report on preliminary experiments exploiting provided data splits for supervised learning.

\subsection{Directional Entailment Measures}
\label{ss:dem}
Note that all directional entailment measures (DEMs) available in the literature have ``pre-embedding'' origins and assume traditional count-based vector spaces \cite{Turney:2010jair,Baroni:2014acl} based on counting word-to-word corpus co-occurrence. Distributional features are typically words co-occurring with the target word in a chosen context (e.g., a window of neighbouring words, a sentence, a document, a dependency-based context). 

This collection of models is grounded on variations of the distributional inclusion hypothesis \cite{Geffet:2005acl}: if $X$ is a semantically narrower term than $Y$, then a significant number of
salient distributional features of $X$ are included in the
feature vector of $Y$ as well. We closely follow the work from Lenci and Benotto \shortcite{Lenci:2012sem} in the presentation. Let $Feat_{X}$ denote the set of distributional features $ft$ for a concept word $X$, and let $w_X(ft)$ refer to the weight of the feature $ft$ for $X$. The most common choices for the weighting function in traditional count-based distributional models are positive variants of pointwise mutual information (PMI) \cite{Bullinaria:2007brm} and local mutual information (LMI) \cite{Evert:2008corpling}.

\paragraph{WeedsPrec ($DEM_1$)}
This DEM quantifies the weighted inclusion of the features of a concept word $X$ within the features of a concept word $Y$ \cite{Weeds:2003emnlp,Weeds:2004coling,Kotlerman:2010nle}:

\vspace{-0.9em}
{\small
\begin{align}
DEM_1(X,Y)=\frac{\sum_{ft \in Feat_{X} \cap Feat_{Y}} w_X(ft)}{\sum_{ft \in Feat_{X}} w_X(ft)}
\end{align}}%

\paragraph{WeedsSim ($DEM_2$)}
It computes the geometrical average of WeedsPrec ($DEM_1$) or some other asymmetric measure (e.g., APinc from Kotlerman et al. \shortcite{Kotlerman:2010nle}) and the symmetric similarity $sim(X,Y)$ between $X$ and $Y$, typically measured by cosine \cite{Weeds:2004coling}, or the Lin measure \cite{Lin:1998acl} as in the balAPinc measure of Kotlerman et al. \shortcite{Kotlerman:2010nle}:

\vspace{-0.9em}
{\small
\begin{align}
DEM_2(X,Y)=DEM_1(X,Y) \cdot sim(X,Y)
\label{eq:dem2}
\end{align}}%

\paragraph{ClarkeDE ($DEM_3$)}
A close variation of $DEM_1$ was proposed by Clarke \shortcite{Clarke:2009gems}:

\vspace{-0.9em}
{\footnotesize
\begin{align}
DEM_3(X,Y)=\frac{\sum_{ft \in Feat_{X} \cap Feat_{Y}} min (w_X(ft),w_Y(ft))}{\sum_{ft \in Feat_{X}} w_X(ft)}
\end{align}}%

\paragraph{InvCL ($DEM_4$)}
A variation of $DEM_3$ was introduced by Lenci and Benotto \shortcite{Lenci:2012sem}. It takes into account both the inclusion of context features of $X$ in context features of $Y$ and non-inclusion of features of $Y$ in features of $X$.\footnote{E.g., if {\em animal} is a hypernym of {\em crocodile},
one expects that (i) a number of context features of {\em crocodile} are also features of {\em animal}, and (ii) that a number of context features of {\em animal} are not context features of {\em crocodile}. As a semantically broader concept, {\em animal} is also found in contexts in which also occur {\em animals} other than {\em crocodiles}.}

\vspace{-0.9em}
{\small
\begin{align}
DEM_4(X,Y)=\sqrt{DEM_3(X,Y) \cdot (1-DEM_3(Y,X))}
\end{align}}%

\subsection{Generality Measures}
\label{ss:gem}
Another related view towards the {\sc type-of} relation is as follows. Given two semantically related words, a key aspect of detecting lexical entailment is the generality of the hypernym compared to the hyponym. For example, {\em bird} is more general than {\em eagle}, having a broader intension and a larger extension. This property has led to the introduction of lexical entailment measures that compare the entropy/semantic content of distributional word representations, under the assumption that a more general term has a higher-entropy distribution \cite{Herbelot:2013acl,Rimell:2014eacl,Santus:2014eacl}. From this group we show the results with the SLQS \cite{Santus:2014eacl} model demonstrating the best performance in prior work.
%SM: From this group, we...
\paragraph{SLQS} It is an entropy-based measure which actually quantifies the specificity/generality level of related terms. First, the top $n$ most associated context features (i.e., typically context words as in the original work of Santus et al. \shortcite{Santus:2014eacl}) are identified (e.g., using positive PMI or LMI); for each identified context feature $cn$, its entropy $H(cn)$ is defined as:

\vspace{-1.1em}
{\small
\begin{align}
H(cn)=-\sum_{i=1}^n P(ft_i|cn) \log_2 P(ft_i|c)
\end{align}}%
where $ft_i$, $i=1,\ldots,n$ is the $i$-th context feature, and $P(ft_i|cn)$ is computed as the ratio of the co-occurrence frequency $(cn,ft_i)$ and the total frequency of $cn$. For each concept word $X$, it is possible to compute its median entropy $E_X$ over the $N$ most associated context features. A higher value $E_X$ implies a higher semantic generality of the concept word $X$. The initial SLQS measure called {\sc SLQS-Basic} is then defined as: 

\vspace{-1.1em}
{\small
\begin{align}
SLQS(X,Y) = 1 - \frac{E_X}{E_Y}
\end{align}}%
This measure may be directly used in standard ungraded LE directionality experiments since $SLQS(X,Y)>0$ implies that X is a type of Y (see Tab.~\ref{tab:bless}). Another variant of SLQS called {\sc SLQS-Sim} is tailored to LE detection experiments: it resembles the $DEM_2$ measure from eq.~\eqref{eq:dem2}, the only difference is that, since SLQS can now produce negative scores, all such scores are set to 0.

\subsection{Visual Generality Measures}
\label{ss:vem}
Kiela et al. \shortcite{Kiela:2015acl} showed that such generality-based measures for ungraded LE need not be linguistic in nature, and proposed a series of visual and multi-modal models for LE directionality and detection. We briefly outline the two best performing ones in their experiments. 

Deselaers and Ferrari \shortcite{Deselaers:2011cvpr} previously showed that sets of images corresponding to terms at higher levels in the WordNet hierarchy have greater visual variability than those at lower levels. They exploit
this tendency using sets of images associated with each concept word as returned by Google's image search. The intuition is that the set of images returned for the broader concept \textit{animal} will consist
of pictures of different kinds of animals, that is, exhibiting greater visual variability and lesser concept specificity; on the other hand, the set of images for {\em bird} will consist of pictures of different birds, while the set for {\em owl} will mostly consist only of images of owls.

The generality of a set of $n$ images for each concept $X$ is then computed. The first model relies on the {\em image dispersion} measure \cite{Kiela:2014acl}. It is the average pairwise cosine distance between all image representations\footnote{As is common practice in multi-modal semantics, each image representation is obtained by extracting the $4096$-dimensional pre-softmax layer from a forward pass in a convolutional neural
network (CNN) \cite{Krizhevsky:2012nips,Simonyan:2015iclr} that has been trained on the ImageNet
classification task using Caffe \cite{Jia:2014mm,Russakovsky:2015ijcv}.} $\{\overrightarrow{i_{X,1}},\ldots,\overrightarrow{i_{X,n}}\}$ for $X$:

\vspace{-0.6em}
{\small
\begin{align}
id(X) = \frac{2}{n(n-1)} \sum_{j<k\leq n} 1 - cos(\overrightarrow{i_{X,j}},\overrightarrow{i_{X,k}})
\label{eq:id}
\end{align}}%
Another similar measure instead of calculating the pairwise distance calculates the distance to the centroid $\overrightarrow{\mu_X}$ of $\{\overrightarrow{i_{X,1}},\ldots,\overrightarrow{i_{X,n}}\}$:

\vspace{-0.5em}
{\small
\begin{align}
cent(X) = \frac{1}{n} \sum_{1\leq j \leq n} 1 - cos(\overrightarrow{i_{X,j}},\overrightarrow{\mu_X})
\label{eq:cent}
\end{align}}%

\paragraph{Final Model} 
The following formula summarises the visual model for ungraded LE directionality and detection which we also test in graded evaluations:

\vspace{-0.5em}
{\small
\begin{align}
s_{\theta} (X,Y) =
\left\{
	\begin{array}{ll}
		1 - \frac{f(X)+\alpha}{f(Y)}  & \mbox{if } cos(\vec{X},\vec{Y}) \geq \theta \\
		0 & \mbox{otherwise }
	\end{array}
\right.
\label{eq:visual}
\end{align}}%
$f$ is one of the functions for image generality given by eq.~\eqref{eq:id} and eq.~\eqref{eq:cent}. The model relying on eq.~\eqref{eq:id} is called {\sc Vis-ID}, while the other is called {\sc Vis-Cent}. $\alpha$ is a tunable threshold which sets a minimum difference in generality for LE identification, driven by the idea that non-LE pairs also have non-identical generality scores.  To avoid false positives where one word is more general but the pair is not semantically related, a second threshold $\theta $ is used, which sets $f$ to zero if the two concepts have low cosine similarity. Finally, $\vec{X}$ and $\vec{Y}$ are representations of concept words used to compute their semantic similarity, e.g., \cite{Turney:2010jair,Kiela:2014emnlp}.

\subsection{Concept Frequency Ratio} 
Concept word frequency ratio (FR) is used as a proxy for lexical generality and it is a surprisingly competitive baseline in the standard (binary) LE evaluation protocols (see Sect.~\ref{ss:evalprot} and later Sect.~\ref{ss:results}) \cite[inter alia]{Weeds:2004coling,Santus:2014eacl,Kiela:2015acl}. The FR model also relies on eq.~\eqref{eq:visual}, the only difference is that $f(X)=freq(X)$, where $freq(X)$ is a simple word frequency count obtained from a large corpus.

\subsection{WordNet-Based Similarity Measures}
\label{ss:wnsim}
A variant of eq.~\eqref{eq:visual} may also be used with any standard WordNet-based similarity measure to quantify the degree of {\sc type-of} relation:

\vspace{-0.5em}
{\small
\begin{align}
s(X,Y) = f_{WN}(X,Y)
\label{eq:wn}
\end{align}}%
% \vspace{-0.5em}
% {\small
% \begin{align}
% s_{\theta} (X,Y) =
% \left\{
% 	\begin{array}{ll}
% 		f_{WN}(X,Y)  & \mbox{if } cos(\vec{X},\vec{Y}) \geq \theta \\
% 		0 & \mbox{otherwise }
% 	\end{array}
% \right.
% \label{eq:wn}
% \end{align}}%
where $f_{WN}(X,Y)$ returns a similarity score based on the WordNet path between two concepts. We use three different standard measures for $f_{WN}$ resulting in three variant WN-based models: 

\noindent (1) {\sc WN-Basic}: $f_{WN}$ returns a score denoting how similar two concepts are, based on the shortest path that connects the concepts in the WN taxonomy.

\noindent (2) {\sc WN-LCh}: Leacock-Chodorow similarity function \shortcite{Leacock:1998wn} returns a score denoting how similar two concepts are, based on their shortest connecting path (as above) and the maximum depth of the taxonomy in which the concepts occur. The score is then $-\log(path/2 \cdot depth)$, where $path$ is the shortest connecting path length and $depth$ the taxonomy depth.

\noindent (3) {\sc WN-WuP}: Wu-Palmer similarity function \cite{Wu:1994acl,Pedersen:2004aaai} returns a score denoting how similar two concepts are, based on the depth of the two concepts in the taxonomy and that of their most specific ancestor node.

Note that all three WN-based similarity measures are not well-suited for graded LE experiments by their design: e.g., they will rank direct co-hyponyms as more similar than distant hypernymy-hyponymy pairs.

\subsection{Order Embeddings}
\label{ss:orderemb}
Following trends in semantic similarity (or graded synonymy computations, see Sect.~\ref{s:graded} again) Vendrov et al. \shortcite{Vendrov:2016iclr} have recently demonstrated that it is possible to construct a {\em vector space} or {\em word embedding} model that specialises in the lexical entailment relation, rather than in the more popular similarity/syonymy relation. The model is then applied in a variety of tasks including ungraded LE detection and directionality.
\begin{figure*}[t]
\centering
\includegraphics[width=0.63\linewidth]{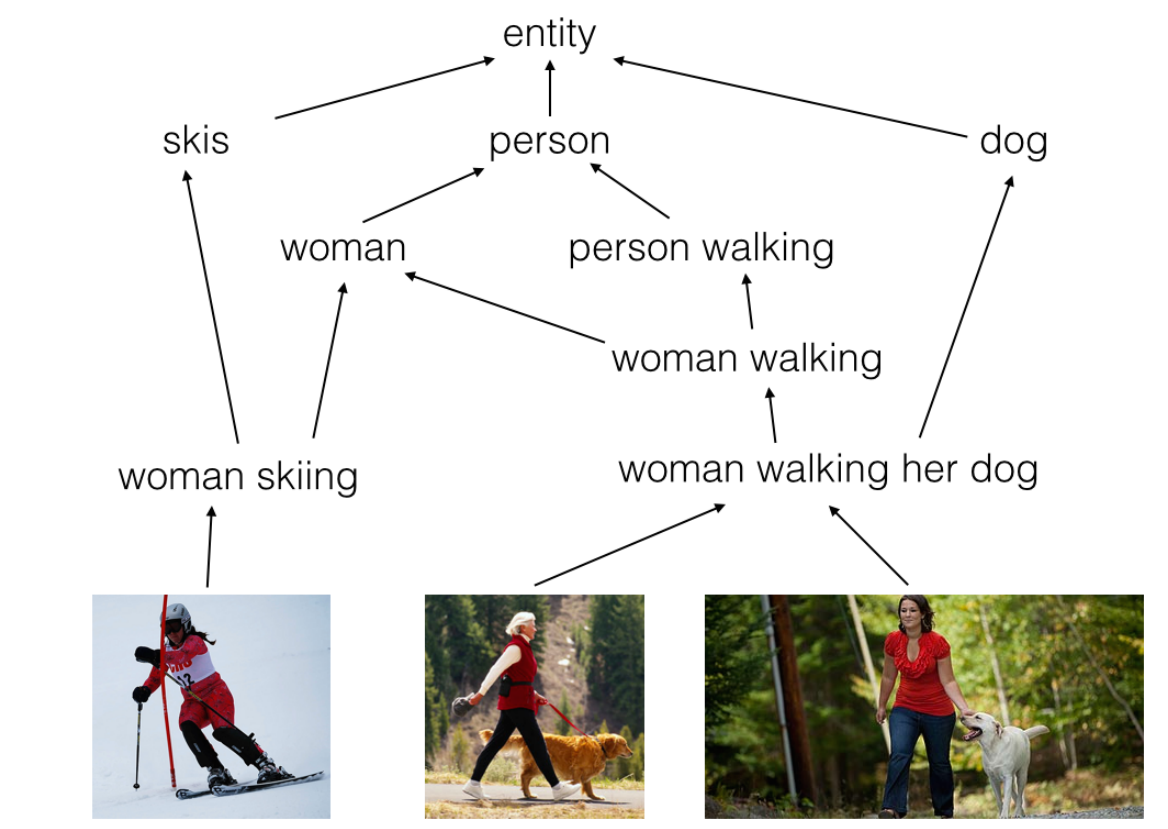}
\vspace{-1.0em}
\caption{A slice of the visual-semantic hierarchy. The toy example taken from Vendrov et al. (2016), inspired by the resource of Young et al. (2014).}
\vspace{-1.3em}
\label{fig:vissem}
\end{figure*}

The order embedding model exploits the partial order structure of a visual-semantic hierarchy (see Fig.~\ref{fig:vissem}) by learning a mapping which is not distance-preserving but order-preserving between the visual-semantic hierarchy and a partial order over the embedding space. It learns a mapping from a partially ordered set $(U,\preceq_{U})$ into a partially ordered embedding space $(V,\preceq_{V})$: the ordering of a pair in $U$ is then based on the ordering in the embedding space. The chosen embedding space is the reversed product order on $\mathbb{R}_{+}^N$, defined by the conjunction of total orders on each coordinate:

\vspace{-1.1em}
{\small
\begin{align}
\vec{X} \preceq \vec{Y} \hspace{1em}\text{ iff } \hspace{1em} \bigwedge_{i=1}^N X_i \geq Y_i
\label{eq:order}
\end{align}}%
for all vectors $\vec{X}$ and $\vec{Y}$ with nonnegative coordinates, where the vectors $\vec{X}$ and $\vec{Y}$ are order embeddings of concept words $X$ and $Y$.\footnote{Smaller coordinates
imply higher position in the partial order. The origin is then the top element of the order, representing the most general concept.} With a slight abuse of notation, $X_i$ refers to the $i$-th coordinate of vector $\vec{X}$, the same for $Y_i$. The ordering criterion, however, is too restrictive to impose as a hard constraint. Therefore, an approximate order-embedding is sought: a mapping which violates the order-embedding condition, imposed as a soft constraint, as little as possible. In particular, the penalty $L$ for an ordered pair $(\vec{X},\vec{Y})$ of points/vectors in $\mathbb{R}_{+}^N$ is defined as:

\vspace{-0.9em}
{\small
\begin{align}
L(\vec{X},\vec{Y}) = ||\max(0,\vec{Y}-\vec{X})||^2
\label{eq:penalty}
\end{align}}%
$L(\vec{X},\vec{Y})=0$ implies that $X \preceq Y$ according to the reversed product order. If the order is not satisfied, the penalty is positive. The model requires a set of positive pairs $PP$ (i.e., true LE pairs) and a set of negative pairs $NP$ for training. Finally, to learn an approximate mapping to an order embedding space, a max-margin loss is used, which encourages positive examples to have zero penalty, and negative examples to have penalty greater than a margin $\gamma$:

\vspace{-0.5em}
{\small
\begin{align}
\sum_{(X,Y) \in PP} L(\vec{X},\vec{Y}) + \sum_{(X',Y') \in NP} \max (0,\gamma-L(\vec{X'},\vec{Y'}))
\label{eq:penalty2}
\end{align}}%
Positive and negative examples are task-dependent. For the standard ungraded LE evaluations, positive pairs for the training set $PP$ are extracted from the WordNet hierarchy. The set $NP$ is obtained by artificially constructing ``corrupted'' pairs \cite{Socher:2013nips}, that is, by replacing one of the two concepts from positive examples with a randomly selected concept. This model is called {\sc OrderEmb}. 

\paragraph{Graded LE with Order Embeddings}
Order embeddings are trained for the binary LE detection task, but not explicitly for the graded LE task. To measure how much one such off-the-shelf order embedding model captures LE on the continuous scale, we test three different distance measures:

\noindent (1) {\sc OrderEmb-Cos}: A standard cosine similarity is used on vector representations.

\noindent (2) {\sc OrderEmb-DistAll}: The sum of the absolute distance between all coordinates of the vectors $\vec{X}$ and $\vec{Y}$ is used as a distance function: 

\vspace{-0.7em}
{\small
\begin{align} 
DistAll(\vec{X},\vec{Y}) = \sum_{i} |Y_{i} - X_{i}|,
\label{eq:orderEmbDistAll}
\end{align}}%
This measure is based on the training penalty defined by eq.~\eqref{eq:penalty}. The idea is that for order embeddings the space is sorted based on the degree of hypernymy/hyponymy violation in each dimension: the absolute coordinate distance may be used as an indicator of the LE strength. 

\noindent (3) {\sc OrderEmb-DistPos}: This variant extends the {\em DistAll} distance by only adding up those coordinates fulfilling the criterion defined in the reversed product order in eq.~\eqref{eq:order}:

\vspace{-0.7em}
{\small
\begin{align}
DistPos(\vec{X},\vec{Y}) = \sum_{i} \begin{cases}
   | Y_{i} - X_{i}| ,& \text{if } X_{i} \geq Y_{i}\\
    0,              & \text{otherwise}
\end{cases}
\label{eq:orderEmbDistPos}
\end{align}}%

% (IV: REMOVED) It would be great to evaluate their hypernymy-based vector space on HyperLex as a sort of intrinsic evaluation (the same way SimLex is considered a very strong intrinsic evaluation for similarity-driven representation models). A small research question in itself is then: how to design a (suitable asymmetric) similarity metric for these order embeddings, e.g., quantifying the degree of hypernym as order violation distance?

\subsection{Standard (``Similarity'') Embeddings}
\label{ss:standard}
A majority of other word embedding models available in the literature target the symmetric relation of semantic relatedness and similarity, and the strength of the similarity relation is modeled by a symmetric similarity measure such as cosine. 

It was shown that human subjects often consider ``closer'' LE pairs quite semantically similar \cite{Geffet:2005acl,Agirre:2009naacl,Hill:2015cl}.\footnote{'Closeness' or hypernymy level for $(X,Y)$ may be measured by the shortest WN path connecting $X$ and $Y$.} For instance, pairs {\em (assignment, task)} or {\em (author, creator)} are judged as strong LE pairs (with average scores 9.33 and 9.30 in HyperLex respectively), they are assigned the labels \texttt{hyp-1} and \texttt{hyp-2} according to WordNet respectively, and are also considered semantically very similar (their SimLex-999 scores are 8.70 and 8.02). In another example, the WordNet \texttt{syn} pairs {\em (foe, enemy)} and {\em (summit, peak)} have graded LE scores of 9.72 and 9.58 in HyperLex. The rationale behind these experiments is then to test to what extent these symmetric models are capable of quantifying the degree of lexical entailment, and to what degree these two relations are interlinked. 

We test the following benchmarking semantic similarity models: (1) Unsupervised models that learn from distributional information in text, including the skip-gram negative-sampling model (\textit{SGNS}) \cite{Mikolov:2013nips} with various contexts ({\sc BOW} = bag of words; {\sc DEPS} = dependency contexts) as described by Levy and Goldberg \shortcite{Levy:2014acl}; (2) Models that rely on linguistic hand-crafted resources or curated knowledge bases. Here, we rely on models using currently holding the peak scores in word similarity tasks: sparse binary vectors built from linguistic resources ({\sc Non-Distributional}, \cite{Faruqui:2015aclnon}), vectors fine-tuned to a paraphrase database ({\sc Paragram}, \cite{Wieting:2015tacl}) further refined using linguistic constraints ({\sc Paragram+CF}, \cite{Mrksic:2016naacl}). Since these models are not the main focus of this work, the reader is referred to the relevant literature for detailed descriptions.

\subsection{Gaussian Embeddings} 
\label{ss:gaussian}
An alternative approach to learning word embeddings was proposed by Vilnis and McCallum \shortcite{Vilnis:2015iclr}. They represent words as Gaussian densities rather than points in the embedding space. Each concept $X$ is represented as a multivariate $K$-dimensional Gaussian parameterised as $\mathcal{N}(\mathbold{\mu}_{X},\mathbold{\sigma}_{X})$, where $\mathbold{\mu}_{X}$ is a $K$-dimensional vector of means, and $\mathbold{\sigma}_{X}$ in the most general case is a $K \times K$ covariance matrix.\footnote{Vilnis and McCallum \shortcite{Vilnis:2015iclr} use a simplification where $\mathbold{\sigma}_{X}$ is represented as a $K$-dimensional vector (so-called {\em diagonal} Gaussian embeddings) or a scalar ({\em spherical} embeddings).}

Word types are embedded into soft regions in space: the intersection of these regions could be straightforwardly used to compute the degree of lexical entailment. This allows a natural representation of hierarchies using e.g. the asymmetric Kullback-Leibler (KL) divergence. KL divergence between Gaussian probability distributions is straightforward to calculate, naturally asymmetric, and has a geometric interpretation as an inclusion between families of ellipses. 

To train the model, they define an energy function that returns a similarity-like measure of the two probabilities. It is possible to train the model to better capture ``standard semantic similarity'' (see Sect.~\ref{ss:standard}) by using expected likelihood (EL) as the energy function. On the other hand, KL divergence is a natural energy function for representing entailment between concepts -- a low KL divergence from $X$ to $Y$ indicates that we can encode $Y$ easily as $X$, implying that $Y$ entails $X$. This can be interpreted as a soft form of inclusion between the level sets of ellipsoids generated by the two Gaussians -- if there is a relatively high expected log-likelihood ratio (negative KL), then most of the mass of $Y$ lies inside $X$.

We refer the reader to the original work \cite{Vilnis:2015iclr,He:2015cikm} for a more detailed description of the idea and actual low-level modelling steps. We evaluate two variants of the model on the graded LE task following \cite{Vilnis:2015iclr}: (i) {\sc Word2Gauss-EL-Cos} and {\sc Word2Gauss-EL-KL} use EL in training, but the former uses cosine between vectors of means as a (symmetric) measure of similarity between concepts, and the latter relies on the (asymmetric) KL divergence between full Gaussians; (ii) {\sc Word2Gauss-KL-Cos} and {\sc Word2Gauss-KL-KL} use KL divergence as the energy function.

\nocite{Young:2014tacl}

\section{Results and Discussion}
\label{s:results}

\subsection{Training Data and Parameters}
\label{ss:parameters}
Since we evaluate a plethora of heterogeneous models and architectures on the graded LE task, we first provide a quick overview of their training setup regarding training data, their parameter settings and other modeling choices.

\paragraph{DEMs and SLQS}
Directional entailment measures $DEM_1$-$DEM_4$ and both SLQS variants (i.e., {\sc SLQS-Basic} and {\sc SLQS-Sim}) are based on the cleaned, tokenised and lowercased Polyglot Wikipedia \cite{AlRfou:2013conll}. We have used two setups for the induction of word representations, the only difference being that in {\em Setup 1} context/feature vectors are extracted from the Polyglot Wiki directly based on bigram co-occurrence counts, while in {\em Setup 2}, these vectors are extracted from the {\sc TypeDM} tensor \cite{Baroni:2010cl} as in the original work of Lenci and Benotto \shortcite{Lenci:2012sem}.\footnote{TypeDM is a variant of the Distributional Memory (DM) framework, where distributional info is represented as a set of weighted word-link-word tuples $\langle\langle w_1,l,w_2 \rangle,\delta \rangle$ where $w_1$ and $w_2$ are word tokens, $l$ is a syntactic co-occurrence link between the words (e.g. a typed dependency link), and $\delta$ is a weight assigned to the tuple (e.g., LMI or PMI).} Both setups use the positive LMI weighting calculated on syntactic co-occurrence links between each word and its context word \cite{Gulordava:2011gems}: $LMI(w_1,w_2) = C(w_1,w_2) * \log_2 \frac{C(w_1,w_2)*Total}{C(w_1) C(w_2)}$,
where $C(w)$ is the unigram count in the Polyglot Wiki for the word $w$, $C(w_1,w_2)$ is the dependency based co-occurrence count of the two tokens $w_1$ and $w_2$., i.e. $(w_1,(dep\_rel,w_2))$, and $Total$ is the number of all such tuples. The Polyglot Wiki was parsed with Universal Dependencies \cite{Nivre:2015ud} as in the work of Vuli\'{c} and Korhonen \shortcite{Vulic:2016acluniversal}.\footnote{We have also experimented with the TypeDM scores directly and negative LMI values. We do not report these results as they are significantly lower than the reported results obtained by the other two setups.} The context vocabulary (i.e., words $w_2$) is restricted to the 10K most frequent words in the Polyglot Wiki. The same two setups were used for the SLQS model. We also use frequency counts collected from the Polyglot Wiki for the frequency ratio model. WordNet-based similarity measures rely on the latest WordNet 3.1 release.

\paragraph{Word Embeddings}
We use 300-dimensional pre-trained order embeddings of Vendrov et al. \shortcite{Vendrov:2016iclr} available online.\footnote{https://github.com/ivendrov/order-embedding} For the detailed description of the training procedure, we refer the reader to the original paper. Gaussian embeddings are trained on the Polyglot Wiki with the vocabulary of the top 200K most frequent single words. We train $300$-dimensional representations using the online tool and default settings suggested by Vilnis and McCallum \shortcite{Vilnis:2015iclr}:\footnote{https://github.com/seomoz/word2gauss} spherical embeddings trained for 200 epochs on a max-margin objective with margin set to 2.

We also use pre-trained standard ``semantic similarity'' word embeddings available online from various sources. $300$-dimensional SGNS-BOW/DEPS vectors are also trained on the Polyglot Wiki: these are the same vectors from \cite{Levy:2014acl}.\footnote{https://levyomer.wordpress.com/2014/04/25/dependency-based-word-embeddings/} $300$-dimensional {\sc Paragram} vectors are the same as in \cite{Wieting:2015tacl}\footnote{http://ttic.uchicago.edu/\textasciitilde wieting/}, while their extension using a retrofitting procedure ({\sc Paragram+CF}) has been made available by Mrk\v{s}i\'{c} et al. \shortcite{Mrksic:2016naacl}.\footnote{https://github.com/nmrksic/counter-fitting} Sparse {\sc Non-Distributional} vectors of Faruqui and Dyer \shortcite{Faruqui:2015aclnon} are also available online.\footnote{https://github.com/mfaruqui/non-distributional}

\subsection{Results}
\label{ss:results}
Due to a wide variety of models and a large space of results used in this work, it is not feasible to present all results at once or provide detailed analyses across all potential dimensions of comparison. Therefore, we have decided to make a gradual selection of the most interesting experiments and results, and stress (what we consider to be) the most important aspects of the HyperLex evaluation set and modeling architectures in our comparisons.

\paragraph{Experiment~I: Ungraded LE Approaches} 
In the first batch of experiments, we evaluate a series of state-of-the-art traditional LE modelling aproaches in the graded LE task on the entire HyperLex evaluation set. The models are described in Sect.~\ref{ss:dem}-Sect.~\ref{ss:wnsim}. A summary of the results is provided in Tab.~\ref{tab:results_dem}. Comparing model scores with the inter-annotator agreements suggests that the graded LE task, although well-defined and understandable by average native speakers, poses a challenge for current ungraded LE models. The absolute difference in scores between human and system performance indicates that there is vast room for improvement in future work. The gap also illustrates the increased difficulty of the graded LE task compared to previous ungraded LE evaluations (see also Exp.~IV). For instance, the best unsupervised LE directionality and detection models from Tab.~\ref{tab:results_dem} reach up over 70\% and up to 90\% in precision scores \cite[inter alia]{Santus:2014eacl,Kiela:2015acl} on BLESS and other datasets discussed in Sect.~\ref{ss:sets}.
\begin{table}[!t]
\begin{center}
\def\arraystretch{0.93}
\begin{footnotesize}
\begin{tabularx}{\linewidth}{l XX}
 \toprule
 {Model} & {\bf Setup 1} & {\bf Setup 2} \\
 \cmidrule(lr){2-3}
 {{\sc FR} ($\alpha=0.02, \theta=0.25$)} & {0.279} & {0.240} \\
  {{\sc FR} ($\alpha=0, \theta=0$)} & {0.268} & {0.265} \\
  \midrule
  {$\text{DEM}_1$} & {0.162} & {0.162} \\
  {$\text{DEM}_2$} & {0.171} & {0.180} \\
  {$\text{DEM}_3$} & {0.150} & {0.150} \\
  {$\text{DEM}_4$} & {0.153} & {0.153} \\
  \midrule
  {\sc SLQS-Basic} & {0.225} & {0.221} \\
  {\sc SLQS-Sim} & {0.228} & {0.226} \\
  \midrule
  {\sc WN-Basic} & {0.207} & {0.207} \\
  {\sc WN-LCh} & {0.214} & {0.214} \\
  {\sc WN-WuP} & {0.234} & {0.234} \\
  \midrule
  {\sc Vis-ID ($\alpha=0.02, \theta=0$)} & {0.203} & {0.203} \\
  {\sc Vis-Cent ($\alpha=0.02, \theta=0$)} & {0.209} & {0.209} \\
  \midrule
  \midrule
  {\sc IAA-1} & {0.854} & {0.854} \\
  {\sc IAA-2} & {0.864} & {0.864} \\
  %\bottomrule
\end{tabularx}
\end{footnotesize}
\end{center}
\vspace{-0.2em}
\caption{Results in the graded LE task over all HyperLex concept pairs obtained by the sets of most prominent LE models available in the literature (see Sect.~\ref{ss:dem}-Sect.~\ref{ss:wnsim}). {\sc Setup 1} and {\sc Setup 2} refer to different training setups for DEMs and SLQS. All results are Spearman's $\rho$ correlation scores. IAA $\rho$ scores are provided to quantify the upper bound for the graded LE task.}
\vspace{-1.2em}
\label{tab:results_dem}
\end{table}

Previous work on ungraded LE evaluation also detected that frequency is a surprisingly competitive baseline in LE detection/directionality experiments \cite{Herbelot:2013acl,Weeds:2014coling,Kiela:2015acl}. This finding stems from an assumption that the informativeness of a concept decreases and generality increases as frequency of the concept increases \cite{Resnik:1995ijcai}. Although the assumption is a rather big simplification \cite{Herbelot:2013acl}, the results based on simple frequency scores in this work further suggest that the FR model may be used as a very competitive baseline in the graded LE task. 

The results also reveal that visual approaches are competitive to purely textual distributional ones. In Tab.~\ref{tab:results_dem}, we have set the parameters according to \cite{Kiela:2015acl}. Varying the $\alpha$ parameter leads to even better results, e.g., the {\sc Vis-ID} model scores $\rho=0.229$ and {\sc Vis-Cent} scores $\rho=0.228$ with $\alpha=1$. This finding supports recent trends in multi-modal semantics and calls for more expressive multi-modal LE models as discussed previously by Kiela et al. \shortcite{Kiela:2015acl}.

To our own surprise, the FR model was the strongest model in this first comparison, while directional measures fall short of all other approaches, although prior work suggested that they are tailored to capture the LE relation in particular. As we do not observe any major difference between two setups for {\sc DEM}s and {\sc SLQS}, all subsequent experiments use Setup 1. The observed strong correlation between frequency and graded LE supports the intuition that prototypical class instances will be more often cited in text, and therefore simply more frequent.

Even WN-based measures do not lead to huge improvements over {\sc DEM}s and fall short of {\sc FR}. Since WordNet lacks annotations pertinent to the idea of graded LE, such simple WN-based measures cannot quantify the actual LE degree. The inclusion of the basic ``semantic relatedness detector'' (as controlled by the parameter $\theta$) does not lead to any significant improvements (e.g., as evident from the comparison of {\sc SLQS-Sim} vs. {\sc SLQS-Basic}, or $DEM_2$ vs. $DEM_1$).

In summary, the large gap between human and system performances along with the FR superiority over more sophisticated LE approaches from prior work unambiguously calls for the next generation of distributional models tailored for graded lexical entailment in particular.

\paragraph{Experiment~II: Word Embeddings} 
In the next experiment, we evaluate a series of state-of-the-art word embedding architectures, covering order embeddings (Sect.~\ref{ss:orderemb}), standard semantic similarity embeddings optimised on SimLex-999 and related word similarity tasks (Sect.~\ref{ss:standard}), and Gaussian embeddings (Sect.~\ref{ss:gaussian}). A summary of the results is provided in Tab.~\ref{tab:results_emb}. The scores again reveal the large gap between the system performance and human ability to consistently judge the graded LE relation. The scores on average are similar to or even lower than scores obtained in Exp.~I. One trivial reason behind the failure is as follows: word embeddings typically apply the cosine similarity in the Euclidean space to measure the distance between $X$ and $Y$. In practice, this leads to the symmetry: $dist(X,Y) = dist(Y,X)$ for each pair $(X,Y)$, which is an undesired model behaviour for graded LE in practice, as corroborated by our analysis of asymmetry in human judgements (see Tab.~\ref{tab:relations} and Tab.~\ref{tab:reversed}). This finding again calls for a new methodology capable of tackling the asymmetry of the graded LE problem in future work.

\begin{table}[!t]
\begin{center}
\def\arraystretch{0.95}
\begin{footnotesize}
\begin{tabularx}{\linewidth}{l XXX}
 \toprule
 {Model} & {\bf All} & {\bf Nouns} & {\bf Verbs}\\
  \cmidrule(lr){2-4}
 {{\sc FR} ($\alpha=0.02, \theta=0.25$)} & {0.279} & {0.283} & {0.239} \\
  {{\sc FR} ($\alpha=0, \theta=0$)}& {0.268} & {0.283} & {0.091} \\
  \midrule
  {\sc SGNS-BOW} (\texttt{win=2}) & {0.167} & {0.148} & {0.289} \\
  {\sc SGNS-DEPS} & {0.205} & {0.182} & {0.352} \\
  \midrule
  {\sc Non-Distributional} & {0.158} & {0.115} & {0.543} \\
  {\sc Paragram} & {0.243} & {0.200} & {0.492} \\
  {\sc Paragram+CF} & {0.320} & {0.267} & {0.629} \\
  \midrule
  {\sc OrderEmb-Cos} & {0.156} & {0.162} & {0.005}\\
  {\sc OrderEmb-DistAll} & {0.180} & {0.180} & {0.130}\\
  {\sc OrderEmb-DistPos} & {0.191} & {0.195} & {0.120}\\
  \midrule
  {\sc Word2Gauss-EL-Cos} & {0.192} & {0.171} & {0.207}\\
  {\sc Word2Gauss-EL-KL} & {0.206} & {0.192} & {0.209}\\
  {\sc Word2Gauss-KL-Cos} & {0.190} & {0.179} & {0.160}\\
  {\sc Word2Gauss-KL-KL} & {0.201} & {0.189} & {0.172}\\
  \midrule
  {\sc IAA-1} & {0.854} & {0.854} & {0.855}\\
  {\sc IAA-2} & {0.864} & {0.864} & {0.862}\\
  %\bottomrule
\end{tabularx}
\end{footnotesize}
\end{center}
\vspace{-0.3em}
\caption{Results (Spearman's $\rho$ correlation scores) in the graded LE task on HyperLex using a selection of state-of-the-art pre-trained word embedding models (see Sect.~\ref{ss:orderemb}-Sect.~\ref{ss:gaussian}). All word embeddings, excluding sparse {\sc Non-Distributional} vectors, are 300-dimensional.}
\vspace{-1.2em}
\label{tab:results_emb}
\end{table}
Dependency-based contexts (SGNS-DEPS) seem to have a slight edge over ordinary bag-of-words contexts (SGNS-BOW) which agrees with findings from prior work on ungraded LE \cite{Roller:2016arxiv,Shwartz:2017eacl}. We observe no clear advantage with {\sc OrderEmb} and {\sc Word2Gauss}, two word embedding models tailored for capturing the hierarchical LE relation naturally in their training objective. We notice slight but encouraging improvements with {\sc OrderEmb} when resorting to more sophisticated distance metrics, e.g., moving from the symmetric straightforward {\sc Cos} measure to {\sc DistPos} with {\sc OrderEmb}, or using {\sc KL} instead of {\sc Cos} with {\sc Word2Gauss}.

As discussed in Sect.~\ref{ss:orderemb}, the off-the-shelf {\sc OrderEmb} model was trained for the binary ungraded LE detection task: its expressiveness for graded LE thus remains limited. One line of future work might utilise the {\sc OrderEmb} framework with a true graded LE objective, and investigate new {\sc OrderEmb}-style representation models fully adapted to the graded LE setting.

\paragraph{Lexical Entailment and Similarity}
Hill, Reichart, and Korhonen \shortcite{Hill:2015cl} report that there is strong correlation between \texttt{hyp-N} word pairs and semantic similarity as judged by human raters. For instance, given the same $[0,10]$ continuous rating scale in SimLex-999, the average similarity score in SimLex-999 for SimLex-999 \texttt{hyp-1} pairs is 6.62, it is 6.19 for \texttt{hyp-2} pairs, and 5.70 for \texttt{hyp-3} and \texttt{hyp-4}. In fact, the only group scoring higher than \texttt{hyp-N} pairs in SimLex-999 are \texttt{syn} pairs with the average score of 7.70. Therefore, we also evaluate state-of-the-art word embedding models obtaining peak scores on SimLex-999, some of them even obtaining scores above the SimLex-999 IAA-1. The rationale is to test whether HyperLex really captures the fine-grained and subtle notion of graded lexical entailment, or the HyperLex annotations were largely driven by decisions at the broader level of semantic similarity.
\begin{figure}[t]
\centering
\includegraphics[width=0.77\linewidth]{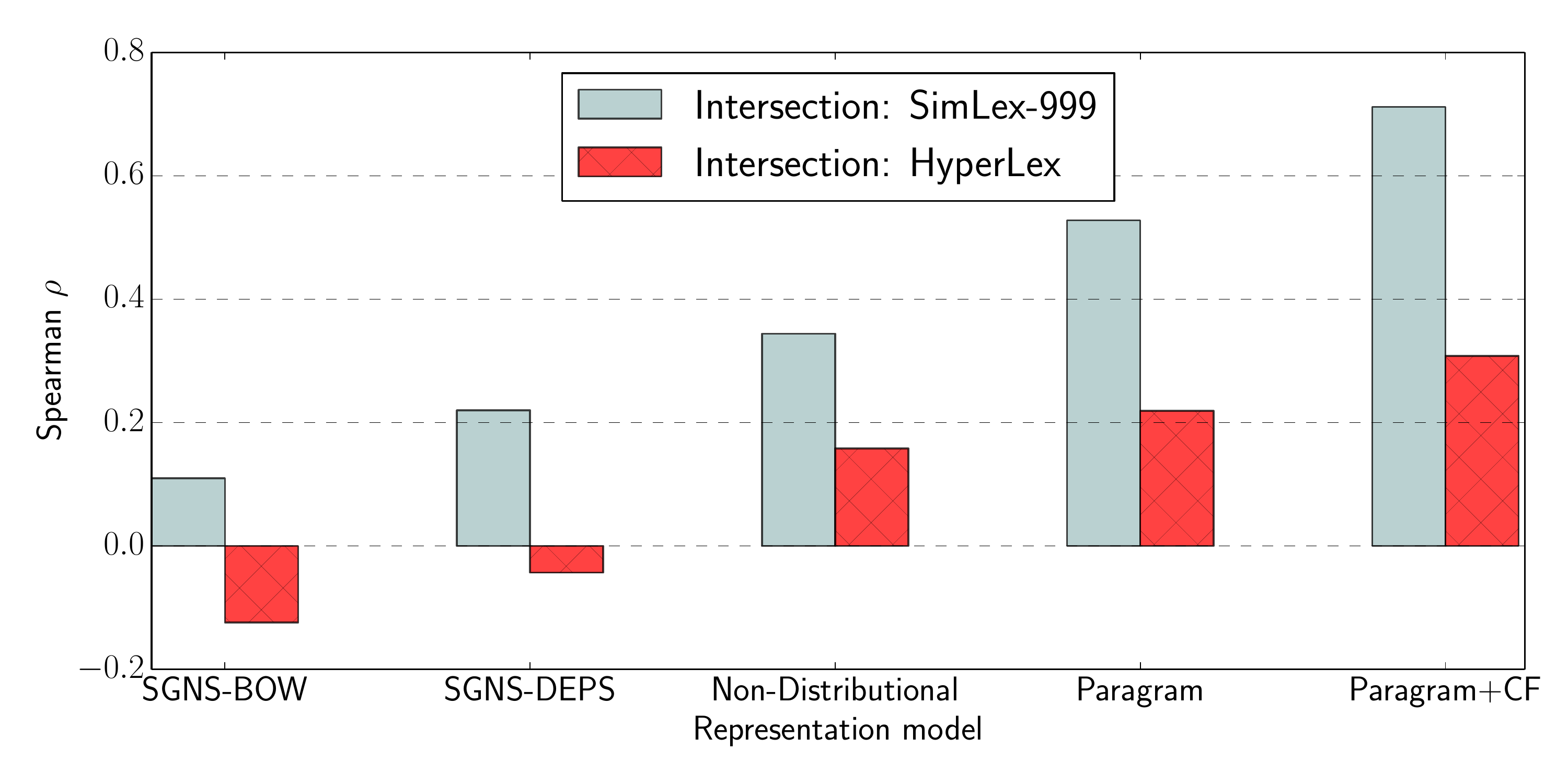}
\vspace{-0.5em}
\caption{Results on the intersection subset of 111 concept pairs annotated both in SimLex-999 (for similarity) and in HyperLex (for graded LE).}
\vspace{-1.2em}
\label{fig:simhyp}
\end{figure}

Another look into Tab.~\ref{tab:results_emb} indicates an evident link between the LE relation and semantic similarity. Positive correlation scores for all models reveal that pairs with high graded LE scores naturally imply some degree of semantic similarity, e.g., {\em author / creator}. However, the scores with similarity-specialised models are much lower than the human performance in the graded LE task, which suggests that they cannot capture intricacies of the task accurately. More importantly, there is a dramatic drop in performance when evaluating exactly the same models in the semantic similarity task (i.e., graded synonymy) on SimLex-999 vs. the graded LE task on HyperLex. For instance, two best performing word embedding models on SimLex-999 are {\sc Paragram} and {\sc Paragram+CF} reaching Spearman's $\rho$ correlation of 0.685 and 0.742, respectively, with SimLex-999 IAA-1 = 0.673, IAA-2 = 0.778. At the same time, the two models score 0.243 and 0.320 on HyperLex respectively, where the increase in scores for {\sc Paragram+CF} may be attributed to its explicit control of antonyms through dictionary-based constraints. 

A similar decrease in scores is observed with other models in our comparisons, e.g., {\sc SGNS-BOW} falls from 0.415 on SimLex-999 to 0.167 on HyperLex. To further examine this effect, we have performed a simple experiment using only the intersection of the two evaluation sets comprising 111 word pairs in total (91 nouns and 20 verbs) for evaluation. The results of selected embedding models on the 111 pairs are shown in Fig.~\ref{fig:simhyp}. It is evident that all state-of-the-art word embedding models are significantly better at capturing semantic similarity.

In summary, the analysis of results with distributed representation models on SimLex-999 and HyperLex suggests that the human understanding of the graded LE relation is not conflated with semantic similarity. Human scores assigned to word pairs in both SimLex-999 and HyperLex reflect truly the nature of the annotated relation: semantic similarity in case of SimLex-999 and graded lexical entailment in case of HyperLex.

\paragraph{Experiment~III: Nouns vs. Verbs} 
In the next experiment, given the theoretical likelihood of variation in model performance across POS categories mentioned in Sect.~\ref{ss:choice}, we assess the differences in results on noun (N) and verb (V) subsets of HyperLex. The results of ``traditional'' LE models (Exp.~I) are provided in Tab.~\ref{tab:results_nv}. Tab.~\ref{tab:results_emb} shows results of word embedding models. IAA scores on both POS subsets are very similar and reasonably high, implying that human raters did not find it more difficult to rate verb pairs. However, we observe differences in performance over the two POS-based HyperLex subsets. First, DEMs obtain much lower scores on the verb subset. It may be attributed to a larger variability of context features for verbs, which also affects the pure distributional models relying on the distributional inclusion hypothesis. WN-based approaches, relying on an external curated knowledge base, do not show the same pattern, with comparable results over pairs of both word classes. Visual models also score better on nouns, which may again be explained by the increased level of abstractness when dealing with verbs. This, in turn, leads to a greater visual variability and incoherence in visual concept representations.
\begin{table}[!t]
\begin{center}
\def\arraystretch{0.93}
\begin{footnotesize}
\begin{tabularx}{\linewidth}{l XX}
 \toprule
 {Model} & {\bf Nouns} & {\bf Verbs} \\
 \cmidrule(lr){2-3}
 {{\sc FR} ($\alpha=0.02, \theta=0.25$)} & {0.283} & {0.239} \\
  {{\sc FR} ($\alpha=0, \theta=0$)} & {0.283} & {0.091} \\
  \midrule
  {$\text{DEM}_1$} & {0.180} & {0.018} \\
  {$\text{DEM}_2$} & {0.170} & {0.047} \\
  {$\text{DEM}_3$} & {0.164} & {0.108} \\
  {$\text{DEM}_4$} & {0.167} & {0.109} \\
  \midrule
  {\sc SLQS-Basic} & {0.224} & {0.247} \\
  {\sc SLQS-Sim} & {0.229} & {0.232} \\
  \midrule
  %|{\sc OrderEmb-cos} & {0.162} & {0.005} \\
  %{\sc OrderEmb-distAll} & {0.180} & {0.130} \\
  %{\sc OrderEmb-distPos} & {0.195} & {0.120} \\
  \midrule
  {\sc WN-Basic} & {0.240} & {0.263} \\
  {\sc WN-LCh} & {0.214} & {0.260} \\
  {\sc WN-WuP} & {0.214} & {0.269} \\
  \midrule
  {\sc Vis-ID ($\alpha=1, \theta=0$)} & {0.253} & {0.137} \\
  {\sc Vis-Cent ($\alpha=1, \theta=0$)} & {0.252} & {0.132} \\
  \midrule
  \midrule
  {\sc IAA-1} & {0.854} & {0.855} \\
  {\sc IAA-2} & {0.864} & {0.862} \\
  %\bottomrule
\end{tabularx}
\end{footnotesize}
\end{center}
\vspace{-0.3em}
\caption{Results in the graded LE task over all HyperLex noun and verb pairs separately. All DEM and SLQS model variants are using Setup 1.}
\vspace{-1.2em}
\label{tab:results_nv}
\end{table}

For word embedding models, we notice that scores for the V subset are significantly higher than for the N subset. To isolate the influence of test set size, we have also repeated experiments with random subsets of the N subset, equal to the V subset in size (453 pairs). We observe the same trend even with such smaller N test sets, leading to a conclusion that difference in results stems from the fundamental difference in how humans perceive nouns and verbs. Human raters seem to associate the LE relation with similarity more frequently in case of verbs, and they do it consistently (based on the IAA scores). We speculate that it is indeed easier for humans to think in terms of semantic taxonomies when dealing with real-world entities (e.g., concrete nouns), than with more abstract events and actions, as expressed by verbs. Another reason could be that, when humans make judgements over verb semantics, syntactic features become more important and implicitly influence the judgements. This effect is supported by the research on automatic acquisition of verb semantics, in which syntactic features have proven particularly important \cite[inter alia]{Kipper:2008lrec,Korhonen:2010rs}.

We leave the underlying causes at the level of speculation. A deeper exploration here is beyond the scope of this work, but this preliminary analysis already highlights how the principal word classes integrated in HyperLex are pertinent to a range of questions concerning distributional, lexical, and cognitive semantics.

\paragraph{Experiment~IV: Ungraded vs. Graded LE}
We also analyse the usefulness of HyperLex as a data set for ungraded LE evaluations and study the differences between graded LE and one ungraded LE task: hypernymy/LE directionality (see Sect.~\ref{ss:evalprot}). First, we have converted a subset of HyperLex into a data set for LE directionality experiments similar to BLESS by retaining only \texttt{hyp-N} pairs from HyperLex (as indicated by WordNet) with the graded LE score $\geq$ 7.0. The subset contains 940 $(X,Y)$ pairs in total (of which 121 pairs are verb pairs), where $Y$ in each pair may be seen as the hypernym. Following that, we run a selection of ungraded LE models from Sect.~\ref{s:experiments} tailored to capture directionality, and compare the scores of the same models in the graded LE task on this HyperLex subset containing ``true hypernymy-hyponymy'' pairs.
\begin{table}[!t]
\begin{center}
\def\arraystretch{0.95}
\begin{footnotesize}
\begin{tabularx}{\linewidth}{l XXX XXX}
 \toprule
 {} & \multicolumn{3}{l}{Directionality} & \multicolumn{3}{l}{Graded LE} \\
 \cmidrule(lr){2-4} \cmidrule(lr){5-7}
 {Model} & {\bf All} & {\bf Nouns} & {\bf Verbs} & {\bf All} & {\bf Nouns} & {\bf Verbs}\\
 \cmidrule(lr){2-4} \cmidrule(lr){5-7}
 {{\sc FR} ($\alpha=0, \theta=0$)} & {0.760} & {0.778} & {0.636} & {0.089} & {0.104} & {0.032} \\
  \cmidrule(lr){2-4} \cmidrule(lr){5-7}
 {$\text{DEM}_1$} & {0.700} & {0.696} & {0.726} & {-0.072} & {-0.102} & {-0.071} \\ 
 {$\text{DEM}_2$} & {0.700} & {0.696} & {0.726} & {-0.070} & {-0.050} & {-0.042} \\ 
 {$\text{DEM}_3$} & {0.696} & {0.684} & {0.777} & {0.036} & {0.063} & {0.115} \\ 
 {$\text{DEM}_4$} & {0.696} & {0.684} & {0.777} & {0.036} & {0.064} & {0.110} \\ 
 \cmidrule(lr){2-4} \cmidrule(lr){5-7}
 {\sc SLQS-Basic} & {0.747} & {0.734} & {0.835} & {0.088} & {0.121} & {-0.036} \\ 
 {\sc SLQS-Sim} & {0.749} & {0.734} & {0.851} & {0.163} & {0.126} & {-0.012} \\
 \cmidrule(lr){2-4} \cmidrule(lr){5-7}
 {\sc OrderEmb} & {0.578} & {0.578} & {0.571} & {0.048} & {0.068} & {0.029} \\
 %\bottomrule
\end{tabularx}
\end{footnotesize}
\end{center}
\vspace{-0.3em}
\caption{Results in the ungraded LE directionality task (precision) using a subset of 940 HyperLex pairs converted to the ungraded directionality data set. Graded LE results (Spearman's $\rho$ correlation) on the same subset are also provided for comparison purposes, using the best model configurations from Tab.~11 and Tab.~12.}
\vspace{-1.8em}
\label{tab:ungraded}
\end{table}

The frequency baseline considers the more frequent concept as the hypernym in the pair. For ${DEM}_1$-${DEM}_4$ models (Sect.~\ref{ss:dem}), the prediction of directionality is based on the asymmetry of the measure: if $DEM_i(X,Y) > DEM_i(Y,X)$, it means that the inclusion of the features of $X$ within the features of $Y$ is higher than the reverse, which in turn implies that $Y$ is the hypernym in the pair. Further, $SLQS(X,Y) > 0$ implies that $Y$ is a semantically more general concept and is therefore the hypernym (see Sect.~\ref{ss:gem}).\footnote{Following the same idea, also discussed in \cite{Lazaridou:2015naacl,Kiela:2015acl}, a concept with a higher word embedding standard deviation or embedding entropy could be considered semantically more general and therefore the hypernym. However, we do not report the scores with word embeddings as they were only slightly better than the random baseline with the precision of 0.5.} With \textsc{OrderEmb}, smaller coordinates mean higher position in the partial order: we compute and compare $DistPos(\vec{X},\vec{Y})$ and $DistPos(\vec{Y},\vec{X})$ scores to find the hypernym. The results provided as binary precision scores are summarised in Tab.~\ref{tab:ungraded}. They reveal that frequency is a strong indicator of directionality, but further improvements, especially for verbs, may be achieved by resorting to asymmetric and generality measures. The reasonably high scores observed in our ungraded directionality experiments are also reported for the detection task in prior work \cite{Shwartz:2017eacl}. The graded LE results on the HyperLex subset are prominently lower than the results with the same models on the entire HyperLex: this shows that fine-grained differences in human ratings in the high end of the graded LE spectrum are even more difficult to capture with current statistical models.

The main message conveyed by the results from Tab.~\ref{tab:ungraded} is that the output from the models built for ungraded LE indeed cannot be used as an estimate of graded LE. In other words, the relative entropy or the measure of distributional inclusion between two concepts can be used to reliably detect which concept is the hypernym in the directionality task, or to distinguish between LE and other relations in the detection task, but it leads to a poor global estimate of the LE strength for graded LE experiments.

\subsection{Supervised Settings: Regression Models} 
\label{ss:regression}
We also conduct preliminary experiments in supervised settings, relying on the random and lexical splits of HyperLex introduced in Sect.~\ref{s:analysis} (see Tab.~\ref{tab:stats}). We experiment with several well-known supervised models from the literature: they typically represent concept pairs as a combination of each concept's embedding vector: concatenation $\vec{X} \oplus \vec{Y}$ \cite{Baroni:2012eacl}, difference $\vec{Y} - \vec{X}$ \cite{Roller:2014coling,Weeds:2014coling,Fu:2014acl}, or element-wise multiplication $\vec{X} \odot \vec{Y}$ \cite{Levy:2015naacl}. Based on state-of-the-art word embeddings such as {\sc SGNS-BOW} or {\sc Paragram}, these methods are easy to apply, and show very good results in ungraded LE tasks \cite{Baroni:2012eacl,Weeds:2014coling,Roller:2014coling}. Using two standardised HyperLex splits, the experimental setup is as follows: we learn a regression model on the {\sc Training} set, optimise parameters (if any) on {\sc Dev}, and test the model's prediction power on {\sc Test}. We experiment with two linear regression models: (1) standard {\em ordinary least squares} ({\sc ols}), and (2) {\em ridge regression} or Tikhonov regularisation ({\sc ridge}) \cite{Myers:1990book}.

%IV (Removed): See Tab.~\ref{tab:stats} for the exact number of items in each of the sets for each split. 

\begin{figure*}[!t]
            \centering
                        \subfigure[Random Split]{
            \includegraphics[width=0.95\linewidth]{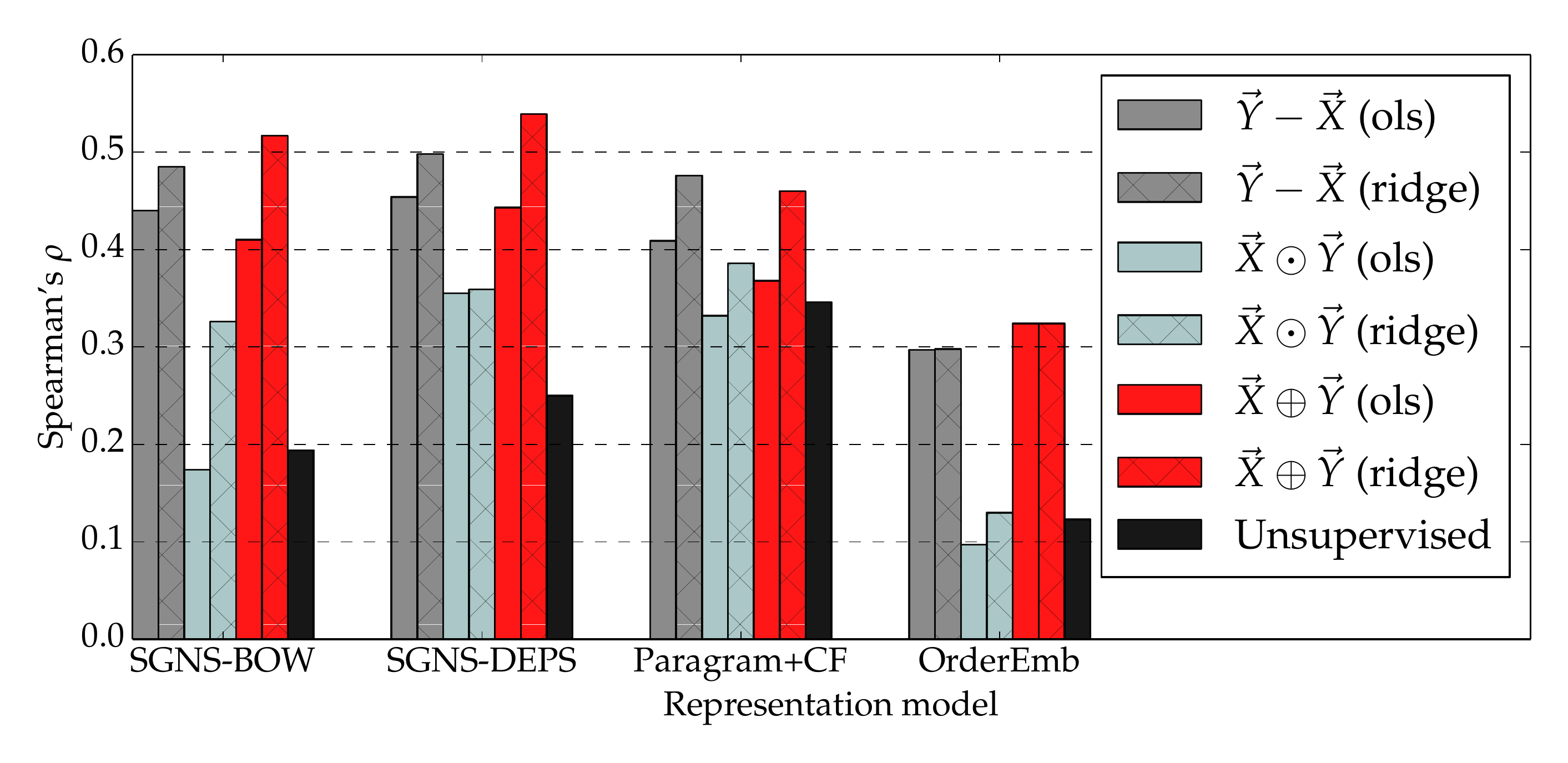}
                                \label{fig:regrandom}
                        }
                        \subfigure[Lexical Split]{
             \includegraphics[width=0.95\linewidth]{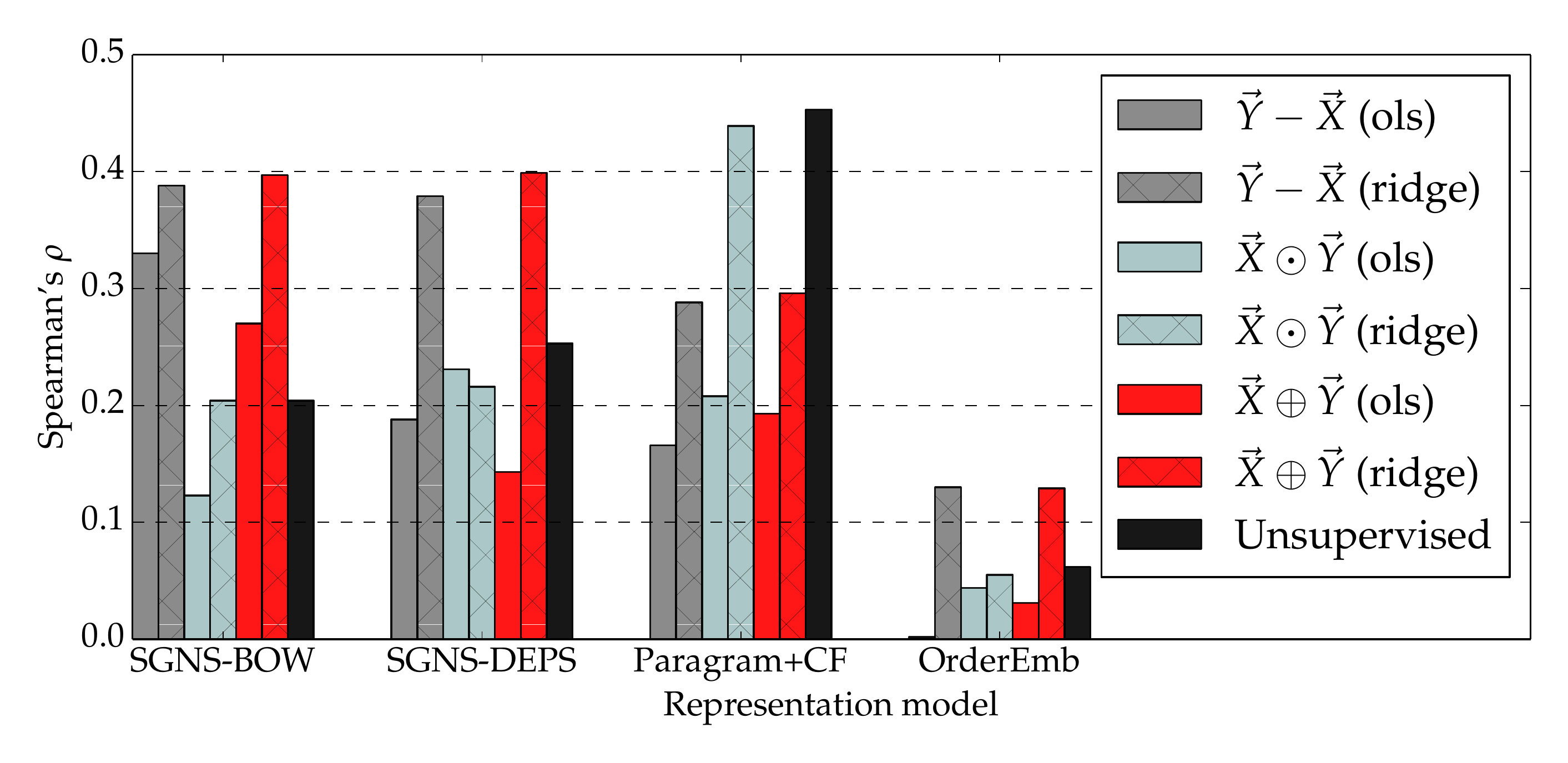}
                                \label{fig:reglexical}
                        }
                        \vspace{-1.1em}
                        \caption{Spearman's $\rho$ correlation scores using two different HyperLex data splits: (a) random and (b) lexical (see Sect.~\ref{s:analysis}). Two linear regression models are used: ordinary least squares ({\sc ols}) and ridge regression ({\sc ridge}), both trained on the training subset of each split, and tested on the test subset. Three typical feature transformations from prior work on LE detection/directionality in supervised settings have been tested: feature vector difference ($\vec{Y}-\vec{X}$), element-wise multiplication ($\vec{X}\odot \vec{Y}$), concatenation ($\vec{X} \oplus \vec{Y}$). We also report baseline $\rho$ correlation scores obtained by simply computing $cos(\vec{X},\vec{Y})$ on each test subset directly, without any model learning ({\sc Unsupervised}). Performance ceilings are 0.849 (IAA-1), 0.862 (IAA-2) for the random split, 0.846 (IAA-1), 0.857 (IAA-2) for the lexical split.}
\vspace{-1.2em}
\end{figure*}
Ridge regression is a variant of least squares regression in which a regularisation term is added to the training objective to favour solutions with certain properties. The regularisation term is the Euclidian L2-norm of the inferred vector of regression coefficients. This term ensures that the regression favors lower coefficients and a smoother solution function, which should provide better generalisation performance than simple {\sc ols} linear regression. The {\sc ridge} objective is to minimise the following:

\vspace{-0.9em}
{\normalsize
\begin{align}
||\vec{a}\mathbf{Q}-\vec{s}||_2^2 + ||\mathbf{\Gamma} \vec{a}||_2^2
\label{eq:ridge}
\end{align}}%
where $\vec{a}$ is the vector of regression coefficients, $\mathbf{Q}$ is a matrix of feature representations for each concept pair $(X,Y)$ obtained using concatenation, difference, or element-wise multiplication. $\vec{s}$ is the vector of graded LE strengths for each concept pair, and $\mathbf{\Gamma}$ is some suitably chosen Tikhonov matrix. We rely on the most common choice: it is a multiple of the identity matrix $\mathbf{\Gamma} = \beta \mathbf{I}$. 

The effect of regularisation is thus varied via the $\beta$ hyperparameter, which is optimised on the {\sc Dev} set. Setting $\beta=0$ reduces the model to the unregularised {\sc ols} solution.

\paragraph{Results}
Following related work, we rely on the selection of state-of-the-art word embedding models to provide feature vectors $\vec{X}$ and $\vec{Y}$. The results of a variety of tested regression models are summarised in Fig.~\ref{fig:regrandom} (random split) and Fig.~\ref{fig:reglexical} (lexical split). As another reference point, we also report results with several unsupervised models on the two smaller test sets in Tab.~\ref{tab:randomlexical}.

\begin{table}[!t]
\begin{center}
\def\arraystretch{0.88}
\begin{footnotesize}
\begin{tabularx}{\linewidth}{l XX}
 \toprule
 {Model} & {\bf Random} & {\bf Lexical}\\
 \cmidrule(lr){2-3}
 {{\sc FR} ($\alpha=0.02, \theta=0.25$)} & {0.299}  & {0.199} \\
 \midrule
  {$\text{DEM}_1$} & {0.212} & {0.188} \\
  {$\text{DEM}_2$} & {0.220} & {0.142} \\
  {$\text{DEM}_3$} & {0.142} & {0.177} \\
  {$\text{DEM}_4$} & {0.145} & {0.178} \\
  \midrule
  {\sc SLQS-Sim} & {0.223} & {0.179} \\
  \midrule
  {\sc WN-Basic} & {0.189} & {0.255} \\
  {\sc WN-WuP} & {0.212} & {0.261} \\
  \midrule
  {\sc Vis-ID ($\alpha=1, \theta=0$)} & {0.203} & {0.201} \\
  {\sc Vis-Cent ($\alpha=1, \theta=0$)} & {0.207} & {0.209} \\
  \midrule
  \midrule
  {\sc IAA-1} & {0.849} & {0.846} \\
  {\sc IAA-2} & {0.862} & {0.857} \\
  \bottomrule
\end{tabularx}
\end{footnotesize}
\end{center}
\vspace{-0.5em}
\caption{Results on the two {\sc Test} sets of HyperLex splits with a selection of unsupervised LE models. Lower scoring model variants are not shown.}
\vspace{-1.4em}
\label{tab:randomlexical}
\end{table}

The IAA scores from Tab.~\ref{tab:randomlexical} again indicate that there is firm agreement between annotators for the two test sets, and that automatic systems still display a large gap to the human performance. The scores on the smaller test sets follow similar patterns as on the entire HyperLex. We see a slight increase of performance for similarity-specialised models (e.g., {\sc WN}-based models or {\sc Paragram+CF}) on the lexical split. We attribute this increase to the larger percentage of verb pairs in the lexical test set, shown to be better modelled with similarity-oriented embeddings in the graded LE task. Verb pairs constitute 17.3\% of the entire random {test} set, the same percentage as in the entire HyperLex, while the number is 26.4\% for the lexical {test} set.

We reassess that supervised distributional methods indeed perform worse on a lexical split \cite{Levy:2015naacl,Shwartz:2016arxiv}. Besides operating with a smaller training set in a lexical split, the finding is also explained by the effect of lexical memorisation with a random split: if high scores are systematically assigned to training pairs \textit{($X_1$, animal)} or \textit{($X_2$, appliance)}, the model will simply memorise that each pair \textit{($Y_1$, animal)} or \textit{($Y_2$, appliance)} should be assigned a high score during inference. The impact of lexical memorisation is illustrated by Tab.~\ref{tab:memorisation} using a sample of concept pairs containing 'prototypical hypernyms' \cite{Roller:2016arxiv} such as \textit{animal}: the regression models assign high scores even to clear negatives such as \textit{(plant, animal)}. However, the effect of lexical memorisation also partially explains the improved performance of all regression models over {\sc Unsupervised} baselines for a random split, as many \textit{($X_t$, animal)} pairs are indeed assigned high scores in the test set. 

\begin{table}[!t]
\begin{center}
\def\arraystretch{0.88}
\begin{footnotesize}
\begin{tabularx}{\linewidth}{X XXX}
 \toprule
 {Concept pair} & {\sc HyperLex} & {\sc ols} & {\sc ridge}\\
 \cmidrule(lr){2-4}
 (plant, animal) & {0.13} & {6.95} & {7.39} \\
 (mammal, animal) & {10.0} & {7.14} & {7.43} \\
 (animal, mammal) & {1.25} & {6.61} & {4.99} \\
 (rib, animal) & {0.35} & {6.94} & {7.08} \\
 (reader, person) & {7.43} & {7.47} & {6.97} \\
 (foot, plant) & {0.42} & {7.86} & {6.05} \\
 (fungus, plant) & {4.75} & {7.94} & {7.51} \\
 (dismiss, go) & {3.97} & {4.22} & {4.29} \\
 (dinner, food) & {4.85} & {9.36} & {8.63} \\
  %\bottomrule
\end{tabularx}
\end{footnotesize}
\end{center}
\vspace{-0.5em}
\caption{The effects of lexical memorisation on the output of regression models when dealing with typical hypernymy concepts higher in the taxonomy (e.g., \textit{animal}, \textit{plant}): {\sc HyperLex} denotes the score assigned to the pair by humans in HyperLex, while {\sc ols} and {\sc ridge} refer to the predicted output of the two tested regression models. We use \textsc{SGNS-DEPS} embeddings with concatenation ($\vec{X} \oplus \vec{Y}$, see Fig.~8), while similar trends are observed with other sets of vectors and feature transformations.}
\vspace{-1.8em}
\label{tab:memorisation}
\end{table}

On the other hand, we also notice that almost all {\sc ols} regression models and a large number of {\sc ridge} models in a lexical split cannot beat unsupervised model variants without any model learning. This suggests that the current state-of-the-art methodology in supervised settings is indeed limited in such scenarios and cannot learn satisfying generalisations regarding the type-of relation between words in training pairs. We suspect that another reason behind strong results with the semantically specialised {\sc Paragram+CF} model in the unsupervised setting for the lexical split is the larger percentage of verbs in the lexical test set as well as explicit handling of antonymy, as mentioned earlier. The model explicitly penalises antonyms through dictionary-based constraints (i.e., pushes them away from each other in the vector space), a property which is desired both for semantic similarity \textit{and} graded LE (see the low scores for the \texttt{ant} relation in Tab.~\ref{tab:relations}).

The variation in results across the tested supervised model variants also indicates that the performance of a regression model is strongly dependent on the actual choice of the underlying representation model, feature transformation, as well as the chosen regression algorithm. First, the results on a random split reveal that the best unsupervised representation model does not necessarily yield the best supervised model, e.g., higher results are observed with {\sc SGNS-DEPS} than with {\sc Paragram} in that setting. {\sc OrderEmb} is by far the weakest model in our comparison. Second, there is no clear winner in the comparison of three different feature representations. While vector difference ($\vec{Y}-\vec{X}$) and concatenation ($\vec{Y}\oplus\vec{X}$) seem to yield higher scores overall for a majority of models, element-wise multiplication obtains highest scores overall in a lexical split with {\sc Paragram} and {\sc Paragram+CF}. The variation clearly suggests that supervised models have to be carefully tuned in order to perform effectively on the graded LE task.

Finally, consistent improvements of {\sc ridge} over {\sc ols} across all splits, models, and feature transformations reveal that the choice of a regression model matters. This preliminary analysis advocates the use of more sophisticated learning algorithms in future work. Another path of research work could investigate how to exploit more training data from resources other than HyperLex to yield improved graded LE models.

%TO DO - CONTINUE (OLS vs RIDGE, LEARNING? ON LEXICAL, NO SINGLE BEST FEATURE REPRESENTATION)

\subsection{Further Discussion: Specialising Semantic Spaces}
\label{ss:further}
Following the growing interest in word representation learning, this work also touches upon the ideas of vector/semantic space specialisation: a desirable property of representation models is their ability to steer their output vector spaces according to explicit linguistic and dictionary knowledge \cite[inter alia]{Yu:2014acl,Wieting:2015tacl,Faruqui:2015naacl,Astudillo:2015acl,Liu:2015acl,Mrksic:2016naacl,Vulic:2017acl}. %For instance, it is possible to specialise semantic spaces to better capture properties of different word classes, and build models that specialise on noun, verb, or adjective similarity \cite{Schwartz:2016naacl,Vulic:2016emnlp}. 
Previous work showed that it is possible to build vector spaces specialised for capturing different lexical relations, e.g., antonymy \cite{Yih:2012emnlp,Ono:2015naacl}, or distinguishing between similarity and relatedness \cite{Kiela:2015emnlp}. Yet, it is to be seen how to build a representation model specialised for the graded LE relation. An analogy with (graded) semantic similarity is appropriate here: it was recently demonstrated that vector space models specialising for similarity and scoring high on SimLex-999 and SimVerb-3500 are able to boost performance of statistical systems in language understanding tasks such as \textit{dialogue state tracking} \cite{Mrksic:2016naacl,Mrksic:2017acl,Vulic:2017acl}. Along the same line, we assume that the specification of what the degree of LE means for each individual pair may also boost performance of statistical end-to-end systems in another language understanding task in future work: natural language inference \cite{Bowman:2015emnlp,Parikh:2016emnlp,Agic:2017arxiv}.

Owing to their adaptability and versatility, we believe that representation architectures inspired by neural networks, e.g., \cite{Mrksic:2016naacl,Vendrov:2016iclr}, are a promising avenue for future modeling work on graded lexical entailment in both unsupervised and supervised settings, despite their low performance on the graded LE task at present.

%TO DO - developments with neural nets (adaptability), relation-specialised representation models with references, exploiting such models in cognitive science applications, etc.

\section{Application Areas: A Quick Overview}
\label{s:application}
The proposed data set should have an immediate impact in the cognitive science research, providing means to analyse the effects of typicality and gradience in concept representations \cite{Hampton:2007cogsci,Decock:2014nous}. Besides this, a variety of other research domains share interest in taxonomic relations, automatic methods for their extraction from text, completion of rich knowledge bases, etc. Here, we provide a quick overview of such application areas for the graded lexical entailment framework and the HyperLex data set.

\paragraph{Natural Language Processing}
As discussed in depth in Sect.~\ref{s:motivation}, lexical entailment is an important linguistic task in its own right \cite{Rimell:2014eacl}. Graded LE introduces a new challenge and a new evaluation protocol for data-driven distributional LE models. In current binary evaluation protocols targeting ungraded LE detection and directionality, even simple methods modeling lexical generality are able to yield very accurate predictions. However, our preliminary analysis in Sect.~\ref{ss:results} demonstrates their fundamental limitations for graded lexical entailment. 

In addition to the use of HyperLex as a new evaluation set, we believe that the introduction of graded LE will have implications on how the distributional hypothesis \cite{Harris:1954word} is exploited in distributional models targeting taxonomic relations in particular \cite[inter alia]{Rubinstein:2015acl,Shwartz:2016arxiv,Roller:2016arxiv}. Further, a tight connection of LE with the broader phrase-/sentence-level task of recognising lexical entailment (RTE) \cite{Dagan:2006pascal,Dagan:2013book} should lead to further implications for text generation \cite{Biran:2013ijcnlp}, metaphor detection \cite{Mohler:2013metaws}, question answering \cite{Sacaleanu:2008coling}, paraphrasing \cite{Androutsopoulos:2010jair}, etc.

\paragraph{Representation Learning}
The previous work on representation learning has mostly focused on the relations of semantic similarity and relatedness, as evidenced by the surge in interest in evaluation of word embeddings on datasets such as SimLex-999, WordSim-353, MEN \cite{Bruni:2014jair}, Rare Words \cite{Luong:2013conll}, etc. This strong focus towards similarity and relatedness means that other fundamental semantic relations such as lexical entailment have been largely overlooked in the representation learning literature. Notable exceptions building word embeddings for LE have appeared only recently (see the work of Vendrov et al. \shortcite{Vendrov:2016iclr} and a short overview in Sect.~\ref{ss:further}), but a comprehensive evaluation resource for intrinsic evaluation of such {\em LE embeddings} is still missing. There is a pressing need to improve, broaden, and introduce new evaluation protocols and datasets for representation learning architectures \cite[inter alia]{Schnabel:2015emnlp,Tsvetkov:2015emnlp,Yaghoobzadeh:2016acl,Faruqui:2016arxiv,Batchkarov:2016repeval}.\footnote{The need for finding better evaluation protocols for representation learning models is further exemplified by the initiative focused on designing better evaluation protocols for semantic representation models (RepEval):\\ \texttt{https://sites.google.com/site/repevalacl16/} \\ \texttt{https://repeval2017.github.io/}} We believe that one immediate application of HyperLex is its use as a comprehensive, wide-coverage large evaluation set for representation-learning architectures focused on the fundamental \textsc{type-of} taxonomic relation.

\paragraph{Data Mining: Extending Knowledge Bases}
Ontologies and knowledge bases such as WordNet, Yago, or DBPedia are useful resources in a variety of applications such as text generation, question answering, information retrieval, or for simply providing structured knowledge to users. Since they typically suffer from incompleteness and a lack of reasoning capability, a strand of research \cite{Snow:2004nips,Suchanek:2007www,Bordes:2011aaai,Socher:2013nips,Lin:2015aaai} attempts to extend existing knowledge bases using patterns or classifiers applied to large text corpora. One of the fundamental relations in all knowledge bases is the {\sc type-of}/{\sc instance-of}/{\sc is-a} LE relation (see Tab.~\ref{tab:kbs} in Sect.~\ref{ss:sets}). HyperLex may be again used straightforwardly as a wide-coverage evaluation set for such knowledge base extension models: it provides an opportunity to evaluate statistical models that tackle the problem of graded LE.

\paragraph{Cognitive Science}
Inspired by theories of prototypicality and graded membership, HyperLex is a repository of human graded LE scores which could be exploited in cognitive linguistics research \cite{Taylor:2003book} and other applications in cognitive science \cite{Gardenfors:2004book,Hampton:2007cogsci}. For instance, reasoning over lexical entailment is related to analogical transfer: transferring information from the past experience (the source domain) to the new situation (the target domain) \cite{Gentner:1983cogsci,Holyoak:2012book}, e.g., seeing an unknown animate object called {\em wampimunk} or {\em huhblub} which resembles a {\em dog}, one is likely to conclude that such {\em huhblubs} are to a large extent types of {\em animals}, although definitely not prototypical instances such as {\em dogs}.
%Again, stuff about prototype theory; the annotations could help shed new light on prototypical/stereotypical instances of a class/category, etc. Anything else for the cognitive science application part? 

\paragraph{Information Search} Graded LE may find application in relational Web search \cite{Cafarella:2006www,Kato:2009cikm,Kopliku:2011jcdl}. A user
of a relational search engine might pose the query: {\em ``List all animals with four legs''} or {\em``List manners of slow movement.''} A system aware of the degree of LE would be better suited to relational search than a simple discrete classifier: the
relational engine could rank the output
list so that more prototypical instances are cited first (e.g., {\em dogs}, {\em cats} or {\em elephants} before {\em huhblubs} or {\em wampimunks}). This has a direct analogy with how standard search engines rank documents or Web pages in descending order of relevance to the user's query. Further, taxonomy keyword search \cite{Song:2011ijcai,Liu:2012kdd,Wu:2012sigmod} is another prominent problem in information search and retrieval where such knowledge of lexical entailment relations may be particularly useful.

\paragraph{Beyond the Horizon: Multi-Modal Modeling} From a high-level perspective, autonomous artificial agents will need to jointly model vision and language in order to parse the visual world and communicate with people. Lexical entailment, textual entailment, and image captioning can be seen as special cases of a partial order over unified visual-semantic hierarchies \cite{Deselaers:2011cvpr,Vendrov:2016iclr}, see also Fig.~\ref{fig:vissem} again. For instance, image captions may be seen as abstractions of images, and they can be expressed at various levels in the hierarchy. The same image may be abstracted as, e.g., {\em A boy and a girl walking their dog}, {\em People walking their dog}, {\em People walking}, {\em A boy, a girl, and a dog}, {\em Children with a dog}, {\em Children with an animal}, etc. Lexical entailment might prove helpful in research on e.g. image captioning \cite{Hodosh:2013jair,Socher:2014tacl,Bernardi:2016jair} or cross-modal information retrieval \cite{Pereira:2014pami} based on such visual-semantic hierarchies, but it is yet to be seen whether the knowledge of gradience and prototypicality may contribute to image captioning systems.

Image generality is closely linked to semantic generality as is evident from recent work \cite{Deselaers:2011cvpr,Kiela:2015acl}. The data set could also be very useful in evaluating models that ground language in the physical world \cite[inter alia]{Silberer:2012emnlp,Silberer:2014acl,Bruni:2014jair}. Future work might also investigate attaching graded LE scores to large hierarchical image databases such as ImageNet \cite{Deng:2009cvpr,Russakovsky:2015ijcv}. %which may further steer the evident trend in joint vision and language modeling.

\section{Conclusion}
\label{s:conclusion}
While the ultimate test of semantic models is their usefulness in downstream applications, the research community is still in need of wide-coverage comprehensive gold standard resources for intrinsic evaluation \cite[inter alia]{Camacho:2015acl,Schnabel:2015emnlp,Tsvetkov:2015emnlp,Hashimoto:2016tacl,Gladkova:2016repeval}. Such resources can measure the general quality of the representations learned by semantic models, prior to their integration in end-to-end systems. We have presented HyperLex, a large wide-coverage gold standard resource for the evaluation of semantic representations targeting the lexical relation of {\em graded} lexical entailment (LE) also known as hypernymy-hyponymy or {\sc type-of} relation, a relation which is fundamental in construction and understanding of concept hierarchies, that is, semantic taxonomies. Given that the problem of concept category membership is central to many cognitive science problems focused on semantic representation, we believe that HyperLex will also find its use in this domain.

The development of HyperLex was principally inspired and motivated by several factors. First, unlike prior work on lexical entailment in NLP, it focuses on the relation of graded or soft lexical entailment at a continuous scale: the relation quantifies the strength of the {\sc type-of} relation between concepts rather than simply making a binary decision as with the ungraded LE variant surveyed in Sect.~\ref{s:motivation}. Graded LE is firmly grounded in cognitive linguistic theory of class prototypes \cite{Rosch:1973:natural,Rosch:1975cognitive} and graded membership \cite{Hampton:2007cogsci}, stating that some concepts are more central to a broader category/class than others (prototypicality) or that some concepts are only within the category to some extent (graded membership). For instance, {\em basketball} is more frequently cited as a prototypical {\em sport} than {\em chess} or {\em wrestling}. One purpose of HyperLex is to examine the effects of prototypicality and graded membership in human judgements, as well as to provide a large repository (i.e., HyperLex contains 2,616 word pairs in total) of concept pairs annotated for graded lexical entailment. A variety of analyses in Sect.~\ref{s:analysis} show that the effects are indeed prominent. 

Second, while existing gold standards measure the ability of models
to capture similarity or relatedness, HyperLex is the first crowdsourced data set with the relation of (graded) lexical entailment as its primary target. As such, it will serve as an invaluable evaluation resource for representation learning architectures tailored for the principal lexical relation, which has plenty of potential applications as indicated in Sect.~\ref{s:application}. Analysis of the HyperLex ratings from more than 600 annotators, native English speakers, showed that subjects can consistently quantify graded LE, and distinguish it from a broader notion of similarity/relatedness and other prominent lexical relations (e.g., cohyponymy, meronymy, antonymy) based on simple non-expert intuitive instructions. This is supported by high inter-annotator agreement scores on the entire data set, as well as on different subsets of HyperLex (e.g., POS categories, WordNet relations).

Third, as we wanted HyperLex to be wide-coverage and representative, the construction process guaranteed that the data set covers concept pairs of different POS categories (nouns and verbs), at different levels of concreteness, and concept pairs standing in different relations according to WordNet. The size and coverage of HyperLex makes it possible to compare the strengths and weaknesses of various representation models via statistically robust analyses on specific word classes, and investigate human judgements in relation to such different properties. The size of HyperLex also enables supervised learning, for which we provide two standard data set splits \cite{Levy:2015naacl,Shwartz:2016arxiv} into training, test, and development subsets.

To dissect the key properties of HyperLex, we conducted a spectrum of experiments and evaluations with most prominent state-of-the-art classes of lexical entailment and embedding models available in the literature. One clear conclusion is that current lexical entailment models optimised for the ungraded LE variant perform very poorly in general. There is clear room under the inter-rating ceiling to guide the development of the next generation of distributional models: the low performance can be partially mitigated by focusing models on the graded LE variant, and developing new and more expressive architectures for LE in future work. Even analyses with a selection of prominent supervised LE models reveal the huge gap between the human and system performance in the graded LE task. Future work thus needs to find a way to conceptualise and encode the graded LE idea into distributional models to tackle the task effectively. Despite their poor performance at present, we believe that a promising step in that direction are neural net inspired approaches to LE proposed recently \cite{Vilnis:2015iclr,Vendrov:2016iclr}, mostly due to their conceptual distinction from other distributional modeling approaches complemented with their modeling adaptability and flexibility. In addition, in order to model hierarchical semantic knowledge more accurately, in future work we may require algorithms that are better suited to fast learning from few examples \cite{Lake:2011cogsci}, and have some flexibility with respect to sense-level distinctions \cite{Reisinger:2010naacl,Neelakantan:2014emnlp,Jauhar:2015naacl,Suster:2016naacl}.

Despite the abundance of reported experiments and analyses in this work, we have only scratched the surface in terms of the possible analyses with HyperLex and use of such models as components of broader phrase- and sentence-level textual entailment systems, as well as in other applications, as quickly surveyed in Sect.~\ref{s:application}. Beyond the preliminary conclusions from these initial analyses, we believe that the benefit of HyperLex will become evident as researchers use it to probe the relationship between architectures, algorithms and representation quality for a wide range of concepts. A better understanding of how to represent the full diversity of concepts (with LE grades attached) in hierarchical semantic networks should in turn yield improved methods for encoding and interpreting the hierarchical semantic knowledge which constitutes much of the important information in language. 

\section*{Acknowledgments}
This work is supported by the ERC Consolidator Grant (no 648909). DK and FH performed their work while they were still at the University of Cambridge.

\bibliographystyle{fullname}
\bibliography{references_hyperlex}

\end{document}